\pdfoutput=1

\documentclass[11pt]{article}

\usepackage[preprint]{acl}

\usepackage{times}
\usepackage{latexsym}

\usepackage[T1]{fontenc}

\usepackage[utf8]{inputenc}

\usepackage{microtype}

\usepackage{inconsolata}

\usepackage{graphicx}
\usepackage{booktabs}
\usepackage{multirow}
\usepackage{multicol}
\usepackage{caption}
\usepackage{subcaption}
\usepackage{bbm}
\usepackage{amsmath}
\usepackage{cleveref}
\usepackage{enumitem}
\usepackage{tipa}
\usepackage{kotex}
\usepackage{xcolor}

\usepackage{makecell}
\usepackage{subcaption}

\title{Happiness is Sharing a Vocabulary: A Study of Transliteration Methods}

\author{Haeji Jung\textsuperscript{1}, Jinju Kim\textsuperscript{2,5}, Kyungjin Kim\textsuperscript{3}, Youjeong Roh\textsuperscript{4}, David R. Mortensen\textsuperscript{5} \\
  \textsuperscript{1}University of British Columbia, Canada \\
  \textsuperscript{2}Sungkyunkwan University, Republic of Korea\\
  \textsuperscript{3}Seoul National University, Republic of Korea\\
  \textsuperscript{4}Electronics and Telecommunications Research Institute, Republic of Korea
  \\
  \textsuperscript{5}Carnegie Mellon University, USA\\
  \small \textbf{Correspondence}: \texttt{haejij@cs.ubc.ca}, \texttt{dmortens@cs.cmu.edu}
  }

\begin{document}
\maketitle
\begin{abstract}
    Transliteration has emerged as a promising means to bridge the gap between various languages in multilingual NLP, showing promising results especially for languages using non-Latin scripts. 
    We investigate the degree to which shared script, overlapping token vocabularies, and shared phonology contribute to performance of multilingual models. To this end, we conduct controlled experiments using three kinds of transliteration (romanization, phonemic transcription, and substitution ciphers) as well as orthography. We evaluate each model on three downstream tasks---named entity recognition (NER), part-of-speech tagging (POS) and natural language inference (NLI)---and find that romanization significantly outperforms other input types in 11 out of 12 evaluation settings, largely consistent with our hypothesis that it is the most effective approach. We further analyze how each factor contributed to the success, and suggest that having longer (subword) tokens shared with pre-trained languages leads to better utilization of the model.
    
\end{abstract}

\section{Introduction}\label{sec:intro}

\begin{figure}
    \centering
    \includegraphics[width=0.8\linewidth]{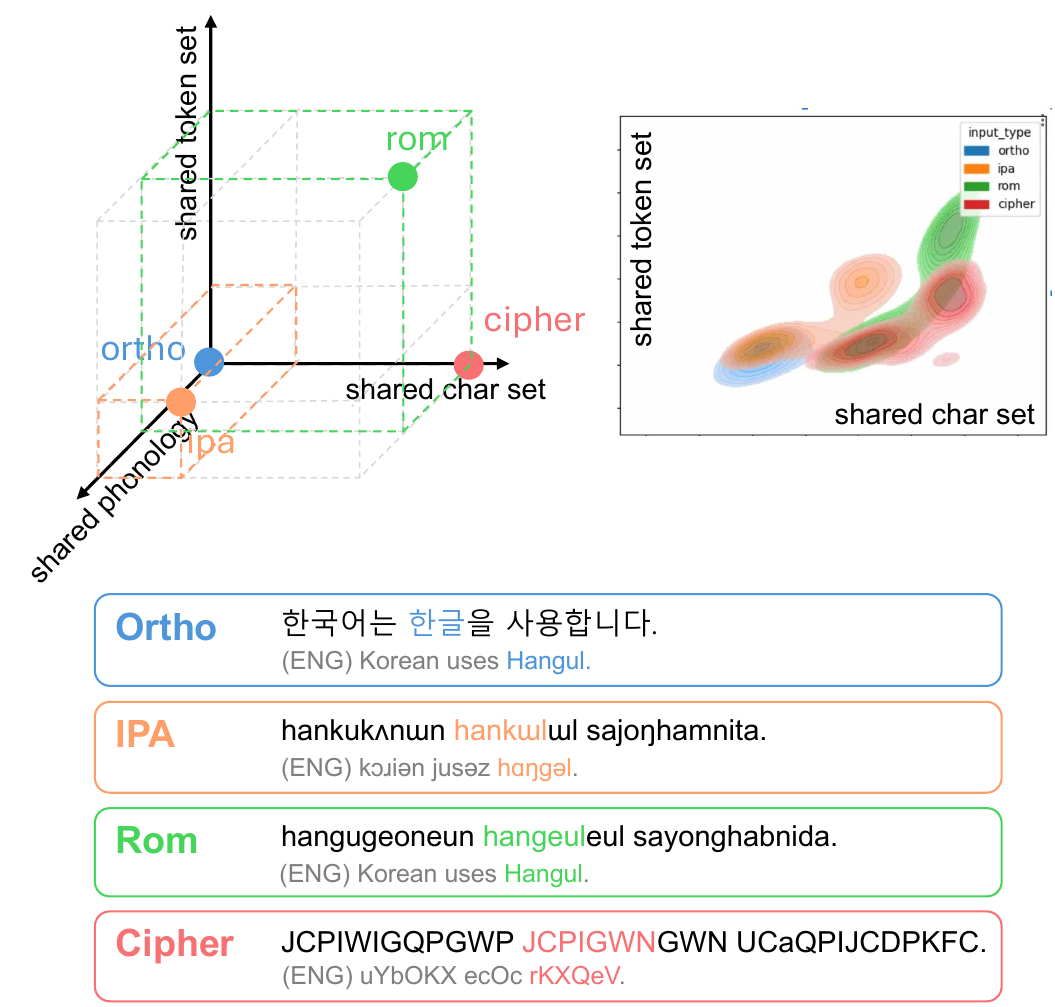}
    \caption{\textit{Top left:} Conceptual visualization of the transliteration analysis schema, positioning input types (Ortho, IPA, Rom, Cipher) based on shared character set, token set, and phonology. \textit{Top right:} KDE plot showing empirical distribution of overlap ratios for each quantifiable component. \textit{Bottom:} Transliteration examples generated with each method.}
    \label{fig:teaser}
\end{figure}
Multilingual language modeling has drawn significant attention from researchers seeking to support 
diverse languages and promote fairness in AI. 
One crucial problem in improving multilingual language modeling is the \textit{script barrier}, a phenomenon in which models struggle to share knowledge between languages written in different scripts, due to mismatched input representations.

Transliteration, which maps a given string to another sequence of characters (e.g., Cyrillic script $\rightarrow$ Latin script) has been explored as a potential solution to the problem of script barriers. In multilingual NLP, transliteration typically maps text to the Latin script or the International Phonetic Alphabet (IPA), giving various languages a shared input representation. Both representations encode linguistic information---specifically phonetic and phonological information---across languages. 
Here, we pose a question: \textit{Is it the shared script itself or the linguistic information encoded in the scripts that helps the models adapt to other languages?}

To investigate this question, we define three key factors in transliteration---(i) shared character set, (ii) shared token set, and (iii) shared phonology of synonymous lexical items---that influence how a model processes and generalizes across languages.
We then designate four input types varying in the degree to which these factors are present: Orthography (Ortho), IPA, Romanized (Rom), and Substitution Ciphered text (Cipher). 
In the context of a script barrier---where two languages are written in different scripts---each input type can be characterized in terms of the three factors as illustrated in \Cref{fig:teaser} and detailed in \Cref{sec:input_types}.

We then conduct controlled experiments to understand how each factor of transliteration contributes to success. First, we pre-train Transformer-based multilingual language models from scratch with each of these four input types, 
and then fine-tune them on target languages---including both seen and unseen languages---to examine how well each input type adapts to new languages. The languages used for pre-training are selected based on our defined language similarity scores as detailed in \Cref{subsec:lang_selection}. We constructed four language sets to account for various scenarios in which language similarity or script familiarity may vary.




We hypothesize that Rom yields the best performance in handling diverse languages with diverse scripts, as it improves input representations across all three dimensions. 
Based on this assumption, IPA is expected to follow, as it enhances two out of three dimensions (shared phonology and tokens) while ciphered text only shares the character set and lacks other shared representations.
Throughout the paper, we evaluate our hypothesis by comparing downstream task performance specifically on unseen languages and analyzing each method in terms of the defined factors, with an emphasis on shared tokens.
We summarize our contribution as follows:
\begin{itemize}[nosep]
    \item We define three factors---shared character set, shared token set, and shared phonology---to explain the effect of transliteration on multilingual inputs.
    \item We conduct controlled pre-training across four language sets and input types, evaluating two downstream tasks to assess how each factor contributes to the success of transliteration.
    \item We analyze vocabulary overlap by token length, revealing how different transliteration methods yield distinct overlap patterns that contribute to cross-lingual adaptation.
\end{itemize}


\section{Related Work}
\label{sec:related_works}

\subsection{Script Barrier and Transliteration}
Script barrier refers to the phenomenon in which multilingual language models struggle to share knowledge between languages written in different scripts (standard orthographies), due to mismatched input representations.
Even massively multilingual models face persistent challenges in handling languages whose scripts were unseen or severely underrepresented during pre-training \citep{de-vries-etal-2022-make,fujinuma-etal-2022-match,muller-etal-2021-unseen,pfeiffer-etal-2021-unks}.

To overcome this limitation, \textit{transliteration} has been explored as an efficient means to improve cross-lingual transfer. Among transliteration methods, \textbf{romanization}---the conversion of non-Latin scripts into the Latin alphabet---has gained popularity due to the dominance of Latin scripts in pre-trained language models \citep{muller-etal-2021-unseen, purkayastha-etal-2023-romanization, husain-etal-2024-romansetu}. 
Another popular method is \textbf{grapheme-to-phoneme} (G2P) conversion, which converts original text (i.e., graphemes) into phonemes represented in International Phonetic Alphabet (IPA). G2P has been applied either by replacing input representations \citep{bharadwaj-etal-2016-phonologically, sohn-etal-2024-zero} or by integrating phonemic information as an auxiliary signal \citep{nguyen-etal-2023-enhancing, nguyen-etal-2025-prompting}.


While these approaches demonstrate transliteration’s practical benefits, relatively little is known about \textit{why} it works.
\citet{moosa-etal-2023-transliteration} examined how transliteration benefits multilingual models by analyzing representational similarity and tokenizer fertility. However, their study mainly focuses on closely related Indic languages, leaving open questions about how transliteration contributes to cross-lingual transfer in a broader, typologically diverse setting.
Similarly, \citet{jung-etal-2024-mitigating} investigated phonemic representations in terms of representational similarity across languages, but without clarifying how transliterated inputs lead to improved cross-lingual performance.

Our work advances these studies by systematically isolating and analyzing the underlying factors---shared character set, shared token set, and shared phonology---that explain when and why transliteration is effective.




\subsection{Vocabulary Overlap in Cross-lingual Transfer}
Lexical overlap has been explored in various works in multilingual NLP. Prior studies have shown that shared lexical or subword units across languages allow models to reuse learned representations, thereby improving transferability \citep{pires-etal-2019-multilingual, chang-etal-2024-multilinguality,limisiewicz-etal-2023-tokenization}.
In contrast, a high proportion of unknown tokens---tokens that the model fails to segment due to their absence from the pre-training corpus---has been shown to hinder transfer and degrade downstream performance \citep{pfeiffer-etal-2021-unks}.


However, \citet{philippy-etal-2023-towards} put former literature together, acknowledging conflicting findings regarding how lexical overlap benefits cross-lingual transfer.
They emphasized the need for task-specific analyses and a more comprehensive understanding of this factor.

In this work, we examine vocabulary overlap to understand its contribution to the success of transliteration. By analyzing overlap by length, we provide deeper insights into how vocabulary overlap shapes multilingual adaptation.

\section{Input Types}
\label{sec:input_types}
\begin{table}[t]
    \centering
    \resizebox{0.9\columnwidth}{!}{\begin{tabular}{cccc}
    \toprule
         & \makecell{Shared \\Char. Set} & \makecell{Shared\\Token Set} & \makecell{Shared \\Phonology} \\ \midrule
         Ortho & -- & -- & -- \\ \midrule
         IPA & $\pm$ & $\pm$ & + \\ \midrule
         Rom & + & + & $\pm$ \\ \midrule
         Cipher & + & -- & -- \\ 
         \bottomrule
    \end{tabular}}
    \caption{Input types characterized by the three factors for transliteration.}
    \label{tab:input_types_factors}
\end{table}


Input types are selected to vary according to three different dimensions:

\begin{itemize}[nosep]
    \item \textbf{Shared Character Set.} Transliteration usually enforces a shared character set across languages. For example, applying romanization to English (e.g., ``hello'' $\rightarrow$ ``hello'') and Korean (e.g., ``안녕하세요'' $\rightarrow$ ``annyeonghaseyo'') produces only Latin characters, which significantly reduces the number of unique characters and patterns that a tokenizer must capture.
    \item \textbf{Shared Token Set.} 
    Transliteration also yield shared subword tokens across languages.
    Here, we specifically distinguish \textit{tokens} from \textit{characters}. By tokens we refer to subword tokens longer than one (more than one character). With English and Korean example again, ``Canada'' $\rightarrow$ ``can\underline{ada}'' and ``캐나다'' $\rightarrow$ ``kaen\underline{ada}'' produce ``\underline{ada}'' as their shared token.
    Since a sequence of characters is more likely to contain semantic meaning than a single character, this distinction is designed to decompose the effect of sharing the surface form and sharing their associated meanings. 
    \item \textbf{Shared Phonology.} 
    Widely used transliteration methods (e.g., G2P and romanization) encode phonological information in their representations. 
    The main motivation for this is that they are more likely to have similar representations for lexical items that sound similar.
    We consider the extent to which they capture the phonology of each language as one of the main factors behind the effectiveness of transliteration, since it enables the recognition of cognates and borrowed vocabulary shared across languages. For example, G2P conversion captures similarities in pronunciation between borrowed words: English ``smartphone'' $\rightarrow$ ``\textipa{smA\*rtfown}'' and Korean ``스마트폰'' $\rightarrow$ ``\textipa{s\textturnm mat\textsuperscript{h}\textturnm p\textsuperscript{h}ʰon}''.

\end{itemize}

To explore these different dimensions of transliteration, we employ four distinct input types: Orthography (Ortho), IPA, Romanized text (Rom), and Substitution Ciphered text (Cipher). How each input type is associated with each factors is summarized in \Cref{tab:input_types_factors}.
The following subsections provide a detailed explanation of each input type and the process of converting written text data (Ortho) into the other input types.

\subsection{Ortho: Standard Orthography}
Different languages use different scripts as their standard orthographies. For example, Korean uses Hangul (한글) and English is written in Latin alphabet. Because models often suffer from significant performance degradation when processing unfamiliar scripts (i.e., \textit{script barrier}), this standard input type serves as a baseline for evaluating the impact of different transliteration methods.

\subsection{IPA: G2P Conversion}
Based on Latin scripts, IPA symbols are designed to represent pronunciations of human language as phonemes. It is therefore considered to be the most accurate method at representing shared phonology. While transliteration into IPA enables some degree of character set sharing, differences in phonemic inventories and phonotactic structures across languages cause each language to use its own distinct set of characters and subword tokens. For example, difference in their phonemic inventories cause `hotel' to be converted into /otel/ in Spanish and /howt\textipa{\textepsilon}l/ in English.

To convert orthographic data into IPA symbols, we use Epitran \cite{mortensen-etal-2018-epitran}. It is a rule-based G2P tool that supports more than a hundred languages, and is widely adopted in transliteration-based methods \citep{bharadwaj-etal-2016-phonologically,chaudhary-etal-2018-adapting,leong-whitenack-2022-phone,sohn-etal-2024-zero,zhu-etal-2024-taste}. We chose a rule-based conversion framework to ensure both consistency and practicality, as such rules are well-established for many languages and the conversion process is much faster.

\subsection{Rom: Romanization} 
Romanization is the process of converting various scripts into Latin letters, enforcing a stricter limit in character sets used across languages.
Additionally, unlike G2P, which often adds language-specificity to languages originally using Latin scripts,
romanization preserves the written form of Latin script.
This allows the resulting text to retain the advantages of a shared script, increasing the likelihood of sharing more subword tokens. 
Since Latin scripts encode sound---though not as precisely as IPA---romanization produces phonologically informed representations for each language. 

For romanization, we employ Uroman \cite{hermjakob-etal-2018-box} which supports conversion of any UTF-8-encoded script into Latin script. 
The tool has been widely used in prior work on romanization and shown to be robust
\citep{amrhein-sennrich-2020-romanization,liu-etal-2024-translico,purkayastha-etal-2023-romanization}.

\subsection{Cipher: Substitution Cipher}
A substitution cipher is a method from cryptography where units of plaintext are replaced with ciphertext according to a predefined rule or key. We apply substitution cipher to the romanized text of each language---with different rules for each---to remove encoded phonological information. While this allows multilingual text to share the same character space as Rom, it no longer encodes cross-lingual phonological patterns and prevents the sharing of meaningful subword tokens across languages. 

We employ a Caesar cipher, a simple substitution encryption technique that shifts each letter in the text by a fixed number of positions in the Latin alphabet. For each language, we assign an integer that determines the shift from the current position of each letter. For example, if English is assigned the integer 4, the word `apple' would be represented as `ettpi', with each letter replaced by the one that is four positions ahead in the alphabet.
Details are provided in \Cref{app:cipher}.


\section{Experimental Setup}
\label{sec:experiments}

\subsection{Language Selection}\label{subsec:lang_selection}
\begin{table}[t]
    \centering
    \resizebox{\linewidth}{!}{
    \begin{tabular}{lllll} 
    \toprule
        Similarity & \makecell{Script \\Diversity} & Language & \makecell{Language\\Family} & Script \\ \midrule
        \multirow{8}{*}{similar} & \multirow{8}{*}{same} & swe &Indo-European&Latn \\ 
        && por&Indo-European&Latn \\
        &&lij&Indo-European&Latn\\
        &&cat&Indo-European&Latn\\
        &&ron&Indo-European&Latn\\
        && spa&Indo-European&Latn \\
        && sqi&Indo-European&Latn \\
        &&fra&Indo-European&Latn \\ \midrule
        \multirow{8}{*}{similar} & \multirow{8}{*}{div} & fra &Indo-European&Latn\\
         && ben&Indo-European&Beng\\
         && hin&Indo-European&Deva\\
         &&hrv&Indo-European&Latn\\
         &&ori&Indo-European&Odia\\
         &&rus&Indo-European&Cyrl\\
         &&srp&Indo-European&Cyrl\\
         &&urd&Indo-European& Arab\\ \midrule
        \multirow{8}{*}{dissimilar} & \multirow{8}{*}{same} & ilo&Austronesian&Latn\\
        && sna&Niger-Congo&Latn\\
        &&lav&Indo-European&Latn\\
        &&uzb&Turkic&Latn\\
        && deu&Indo-European&Latn\\
        &&fin&Uralic&Latn\\
        && som&Afroasiatic&Latn\\
        && swa&Niger-Congo&Latn \\ \midrule
        \multirow{8}{*}{dissimilar} & \multirow{8}{*}{div} & amh&Afroasiatic&Ethi\\
         && ben&Indo-European&Beng\\
         &&tel&Dravidian&Telu\\
         &&fra&Indo-European&Latn\\
         && tha&Tai-Kaidai&Thai\\
         &&kat&Caucasian&Geor\\
         &&kor&Koreanic&Hang\\
         &&mya&Sino-Tibetan&Mymr \\ 
    \bottomrule
    \end{tabular}
    }
    \caption{Languages selected for each language set.}
    \label{tab:language_sets}
\end{table}
To examine how different input types impact multilingual adaptation, we selected languages to form four language sets: (i) typologically similar languages using the same script (sim-same), (ii) similar languages using diverse scripts (sim-div), (iii) dissimilar languages using the same script (dissim-same), and (iv) dissimilar languages using diverse scripts (dissim-div). 
Similar to \citet{chang-etal-2024-multilinguality}, we utilized lang2vec \cite{littell-etal-2017-uriel}\footnote{Utilizing https://github.com/antonisa/lang2vec} to compute language similarity. We extracted syntactic, geographic, and genetic feature vectors from lang2vec to obtain cosine similarities, and also defined lexical similarity based on word overlap ratio between training corpora of each language\footnote{Words are segmented by white spaces.}.
We assigned eight languages to each set (see \Cref{tab:language_sets}) by sampling languages based on the aggregated scores. More details are provided in \Cref{app:language_selection}. 


\subsection{Vocabulary Overlap Ratio}\label{subsec:vocab_overlap}
As discussed in \Cref{sec:input_types}, different transliteration methods are likely to produce different patterns of token overlap.
We therefore focus on analyzing overlaps between target language and pretraining languages to understand how each transliteration method benefits.
Throughout the analysis, we compute the overlap ratio as follows:

\begin{equation}
    \mathrm{OverlapRatio}(l_t, L_s) = \max\limits_{l\in L_s} \frac{|S_{l} \cap S_{l_t}|}{|S_{l_t}|}
\end{equation}
where $L_s$ is a set of pre-trained languages, $l_t$ the target language, and $S_l$ the set of tokens from the language $l$. We suppose all sets $S$ that are used to compute overlap ratio is based on the same tokenizer. The overlap ratio of $l_t$ is define as the maximum overlap achieved between $l_t$ and and any pre-trained language.

We further break down the overlaps by length to examine which lengths contribute most to the results through shared tokens. The overlap ratio for tokens of length $m$ is computed as

    \begin{equation}\label{eq:overlap_by_length}
    \begin{aligned}
        \frac{|\{x \in S_{l_s} \cap S_{l_t} \mid \mathrm{len}(x)=m\}|}{|S_{l_t}|}\\
        \text{where\quad}    
        l_s=\operatorname*{arg\,max}_{l \in L_s}\frac{|S_{l} \cap S_{l_t}|}{|S_{l_t}|} \;.
    \end{aligned}
    \end{equation}
We compute the overlap ratio using the pre-trained language that exhibits the greatest overlap.
Counting token overlaps across all pre-trained languages would, in most cases, cover nearly all tokens from the target language, since the tokenizer is trained on those languages.
For a more meaningful analysis, we therefore focus on the single pre-trained language that the target language is most likely to leverage for our main analysis. Analyses based on overlaps with all pre-trained languages are provided in the \Cref{appx:diff_compute}.



\subsection{Datasets}
For pre-training, we utilize sampled version of a preprocessed Wikipedia corpus from Hugging Face.\footnote{https://huggingface.co/datasets/wikimedia/wikipedia}
For downstream tasks, we utilized WikiAnn \cite{pan-etal-2017-cross, rahimi-etal-2019-massively} dataset for NER, Universal Dependencies (v2.17) \cite{nivre-etal-2020-universal} for POS, and XNLI \cite{conneau-etal-2018-xnli} for sentence classification (NLI) task. More details on preprocessing and dataset statistics can be found in \Cref{app:dataset}.
In order to train the model with different input types, we converted all datasets into each input type using the aforementioned tools.

\subsection{Model Training}
To investigate the impact of different input types, we pre-train 16 models from scratch using four input types and four language sets. We avoid using publicly available pre-trained models to ensure a controlled experimental setup, as most such models are trained on orthographic data, which would confound the analysis of transliteration effects.

For our controlled experiments, we first trained a SentencePiece tokenizer (BPE) for each model with fixed vocabulary size of 30K for all tokenizers. For model architecture, we employed Transformer encoder architecture as in XLM-R \citep{conneau-etal-2020-unsupervised}---a widely known model for its compelling multilingual performances in various tasks.
As with XLM-R, we follow the training regime used for RoBERTa \cite{roberta} with masked language modeling on each multilingual corpus. After pre-training, we fine-tune each model on target language dataset to evaluate its downstream performance. 
For details on the model configurations and training, refer to \Cref{app:model} and \Cref{app:training}.



\section{Results} \label{sec:results}

\begin{table*}[h]
    \centering
    \resizebox{0.9\textwidth}{!}{\begin{tabular}{llrrrrrrrrr}
        \toprule
         \multirow{2}{*}{\makecell{Trained\\Lang. Set}} & \multirow{2}{*}{\makecell{Input\\Type}} & \multicolumn{3}{c}{NER} & \multicolumn{3}{c}{POS} & \multicolumn{3}{c}{NLI} \\\cmidrule{3-11}
         &&\multicolumn{1}{c}{All}&\multicolumn{1}{c}{Seen} & \multicolumn{1}{c}{Unseen}&\multicolumn{1}{c}{All} &\multicolumn{1}{c}{Seen} & \multicolumn{1}{c}{Unseen}&\multicolumn{1}{c}{All} &\multicolumn{1}{c}{Seen} & \multicolumn{1}{c}{Unseen}  \\\midrule
         \multirow{5}{*}{sim-same}& \# Lang. & 28 & 8&20 & 19& 7 & 12 &11 & 2&9\\ \cmidrule{2-11}
         &Ortho &0.7141& \textbf{0.8466} & 0.6611&0.8403&0.9513&0.7755&0.5847&\textbf{0.6768}&0.5642 \\
         &IPA&0.7168&0.8085&0.6801&0.8959&0.9386&0.8709&0.6057&0.6605&0.5936\\
         &Rom&\textbf{0.7589}&0.8395&\textbf{0.7267}&\textbf{0.9088}&\textbf{0.9499}&\textbf{0.8848}&\textbf{0.6276}&0.6755&\textbf{0.6170}\\
         &Cipher&0.7210&0.8173&0.6824&0.9042&0.9463&0.8796&0.6133&0.6658&0.6017\\\midrule
         \multirow{5}{*}{sim-div}&\# Lang.&28&8&20&19&5&14&11&4&7\\\cmidrule{2-11}
         &Ortho&0.6917&0.8409&0.6321&0.8150&0.9501&0.7667&0.6051&0.6242&0.5942\\
         &IPA&0.7202&0.8239&0.6787&0.8942&0.9446&0.8762&0.6171&0.6192&0.6160\\
         &Rom&\textbf{0.7522}&\textbf{0.8451}&\textbf{0.7151}&\textbf{0.9020}&\textbf{0.9511}&\textbf{0.8844}&\textbf{0.6248}&\textbf{0.6273}&\textbf{0.6233}\\
         &Cipher&0.7200&0.8270&0.6772&0.8953&0.9473&0.8767&0.6128&0.6104&0.6142\\\midrule
         \multirow{5}{*}{dissim-same}&\# Lang.&28&7&21&19&4&15&11&2&9\\\cmidrule{2-11}
         &Ortho&0.6935&0.7860&0.6626&0.8189&\textbf{0.8950}&0.7986&0.6000&0.6243&0.5947\\
         &IPA&0.7534&0.7732&0.7468&0.8737&0.8673&0.8754&0.6093&\textbf{0.6307}&0.6045\\
         &Rom&0.7455&\textbf{0.7981}&0.7280&\textbf{0.8912}&\textbf{0.8950}&\textbf{0.8902}&\textbf{0.6155}&0.6195&\textbf{0.6146}\\
         &Cipher&\textbf{0.7592}&0.7725&\textbf{0.7547}&0.8832&0.8913&0.8810&0.5995&0.6296&0.5928\\\midrule
         \multirow{5}{*}{dissim-div}&\# Lang.&28&8&20&19&5&14&11&2&9\\\cmidrule{2-11}
         &Ortho&0.7436&0.7402&0.7450&0.8399&0.8819&0.8249&0.5670&\textbf{0.6481}&0.5490\\
         &IPA&0.7524&0.7524&0.7524&0.8991&0.8925&0.9014&0.6220&0.6039&0.6260\\
         &Rom&\textbf{0.7748}&\textbf{0.7538}&\textbf{0.7832}&\textbf{0.9059}&\textbf{0.8928}&\textbf{0.9106}&\textbf{0.6327}&0.6172&\textbf{0.6361}\\
         &Cipher&0.7502&0.7518&0.7496&0.8987&0.8896	&0.9019&0.6268&0.6176&0.6288\\\bottomrule
         
    \end{tabular}}
    \caption{Downstream task performances for each model---F1 scores for NER and POS, and accuracy scores for NLI. Average scores for all, seen, and unseen languages are reported for each task. \textbf{Bold} indicates best performing input type per language set.}
    \label{tab:ner}
\end{table*}


\begin{figure*}[]
    \centering
    \begin{subfigure}[]{0.25\textwidth}
        \centering
        \includegraphics[height=1.2in]{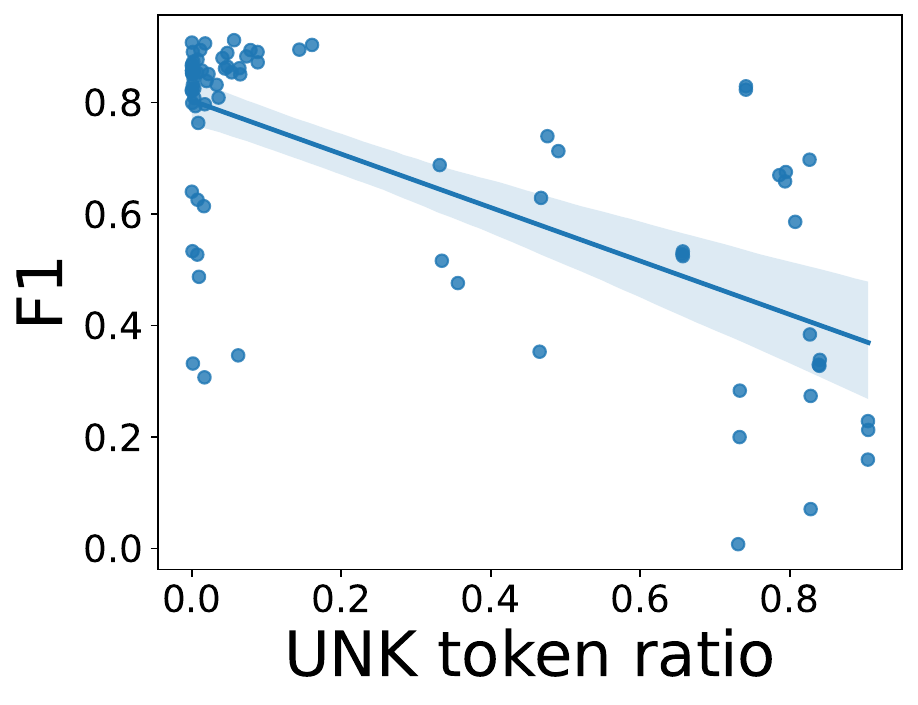}
        \caption{\label{fig:unk_ratio_f1_corr}} 
    \end{subfigure}%
    \begin{subfigure}[]{0.8\textwidth}
        \centering
        \includegraphics[height=1.3in]{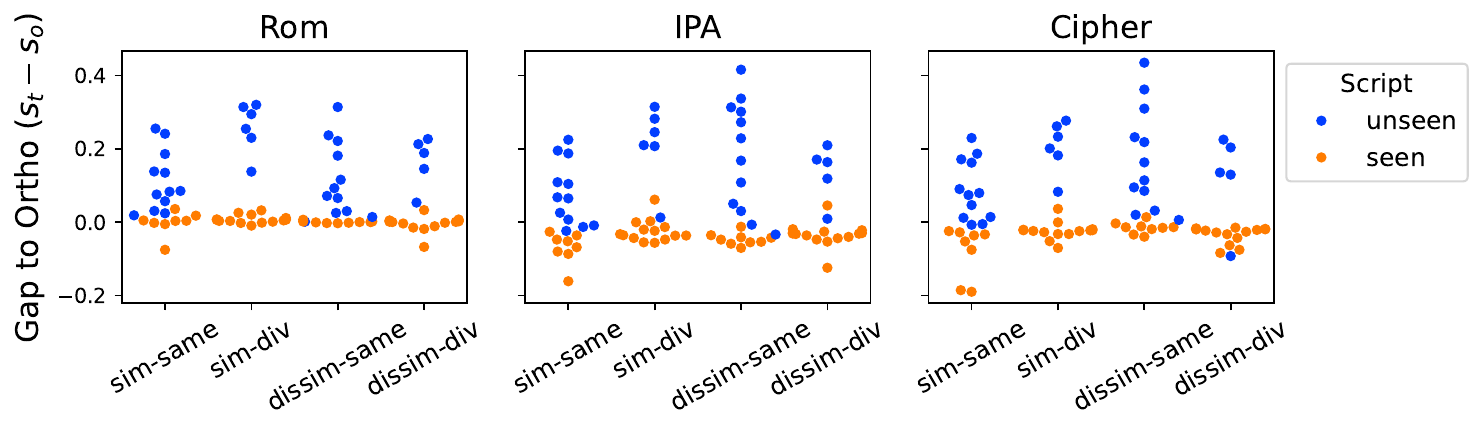}
        \caption{\label{fig:transliteration_gain_script_seen}}
    \end{subfigure}
    \caption{(a) Negative correlation between UNK token ratio and F1 score. (b) Performance gains of transliterated input types compared to orthography-based models, where $s_t$ and $s_o$ denote scores with transliterated and orthographic inputs, respectively. Performance gains appear primarily in target languages whose original scripts are unseen during pre-training.
    }
    
    
\end{figure*}

\Cref{tab:ner} presents average scores across target languages for the two downstream tasks---F1 scores for NER and average accuracies on XNLI---for seen and unseen languages\footnote{Unseen languages refer to languages not included in pre-training of each model.}. $P$-values obtained from paired $t$-tests on F1 scores across input types can be found in \Cref{app:ttest-pvalue}. 

\paragraph{NER Results}
We observe that Rom outperforms other input types on average for unseen languages, which simulate the script barrier. Even for seen languages, Rom performs comparably to Ortho.
To further verify the significance of the results,
we conduct paired $t$-tests to examine the statistical significance of differences between input types.
We find that Rom yields a significant advantage over all other input types for unseen languages ($p < 0.05$), whereas Ortho performs worse than all other input types ($p < 0.05$).
Interestingly, IPA and Cipher do not differ significantly,
despite Cipher encoding no shared linguistic information across languages.

\paragraph{POS Results}
Results on the POS task exhibit similar trends to those observed for NER, with Rom outperforming all other input types even on seen languages, and Ortho significantly underperforming on unseen languages. We additionally report the results of paired $t$-tests in \Cref{app:ttest-pvalue}.

\paragraph{XNLI Results}
For XNLI, we evaluated on 11 languages, most of which were unseen during pre-training. Similar to NER, transliteration does not improve performance on seen languages, with no statistically significant differences between input types. However, for unseen languages, all transliterated input types outperform Ortho, with Rom consistently showing the strongest performance across all language sets. Paired $t$-tests examining the differences between input types support these findings (See \Cref{app:ttest-pvalue}).

\section{Analysis and Discussion}\label{sec:analysis}
In this section, we analyze how each factor of transliteration---shared characters, shared (subword) tokens, and shared phonology---is associated with the downstream performance. Given that transliteration benefits unseen languages rather than seen ones, our analysis focuses on unseen languages unless otherwise noted.


\subsection{Shared Character Set: Overcoming UNK Tokens}\label{subsec:shared_char_set}




\Cref{fig:unk_ratio_f1_corr} shows that the script barrier is closely related to the proportion of unknown (UNK) tokens, which we compute as the number of UNK tokens divided by the total number of tokens in the evaluation data.
A UNK token occurs when the tokenizer fails to segment a sequence of characters using its learned vocabulary, and high UNK ratios arise primarily in languages written in unseen scripts, where the tokenizer has no prior exposure to the characters.

\begin{figure}[]
    \centering
    \includegraphics[width=0.75\linewidth]{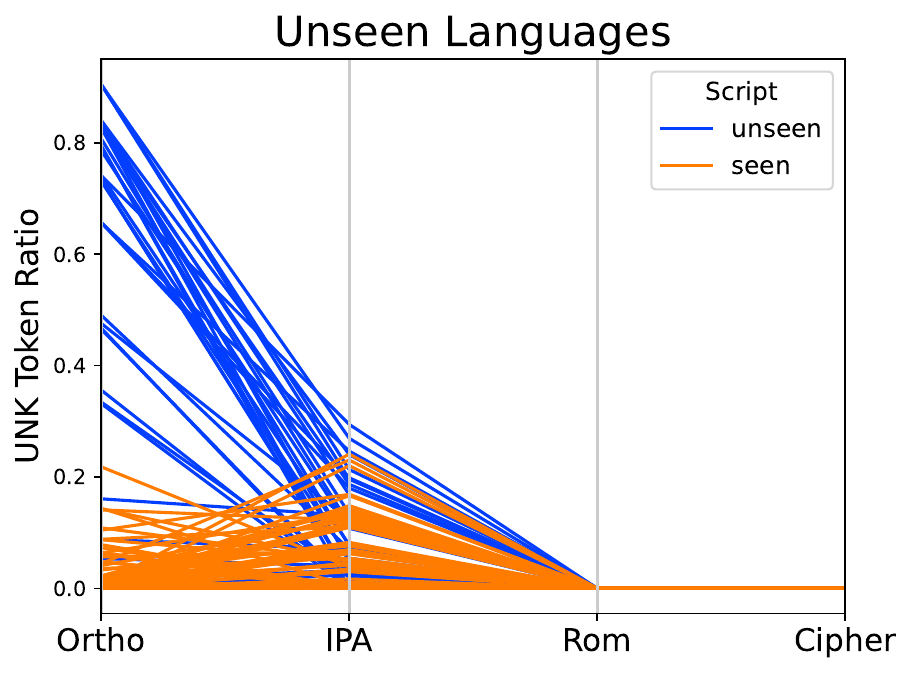}
    \caption{Unknown (UNK) token ratio for unseen languages across different input types.}
    \label{fig:reduce_unk}
\end{figure}

Consequently, applying transliteration and sharing characters across languages greatly reduces UNK ratio in unseen languages (See \Cref{fig:reduce_unk}), leading to clear performance gains over Ortho as shown in \Cref{fig:transliteration_gain_script_seen}.
To this end, \textbf{shared character set} serves as the initial means by which transliteration overcomes the script barrier, reducing the proportion of UNK tokens.
Results from Cipher further support this interpretation: despite containing no semantic or linguistic information shared across languages, its simple character sharing yields notable performance gains for languages written in unseen scripts compared to Ortho.


\subsection{Sharing Longer Token Matters}\label{subsec:shared_token_set}

While reducing the UNK ratio substantially mitigates the script barrier for languages written in unseen scripts, the correlation between UNK ratio and downstream performance becomes less pronounced across different transliteration methods---IPA, Rom, and Cipher.
To further investigate what drives these remaining differences, we analyze token overlaps across languages under each transliteration method, examining how lengths of overlapping tokens relate to downstream performance. 



\begin{figure}
    \centering
    \includegraphics[width=\linewidth]{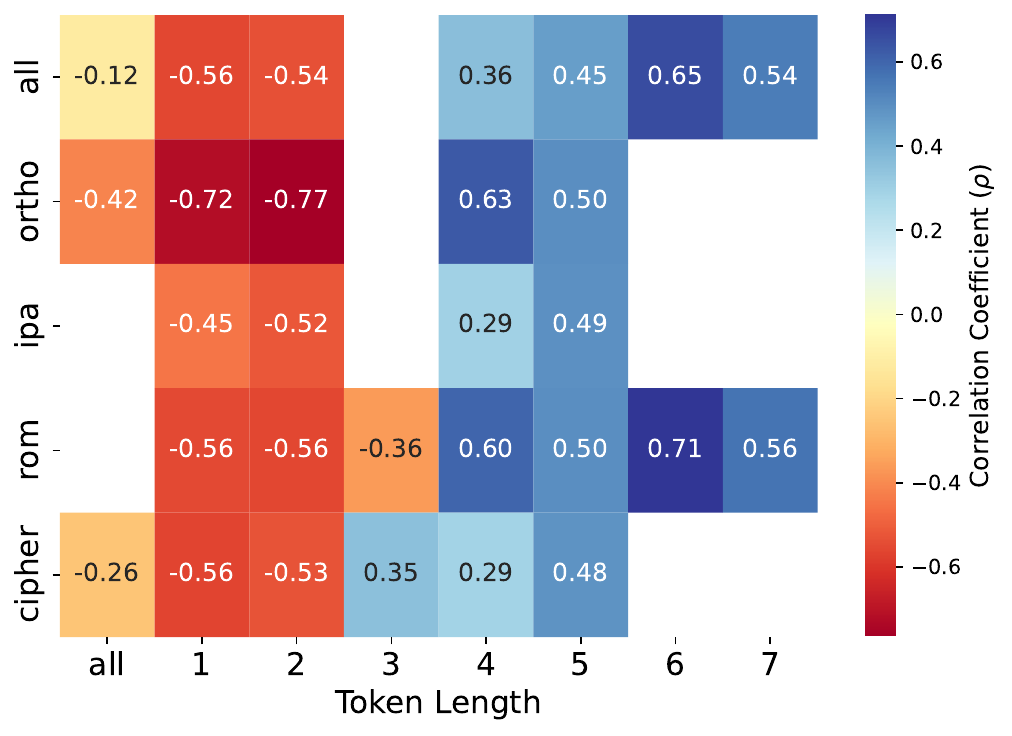}
    \caption{Correlation between token overlap ratio by length and downstream performance. Correlations with $p > 0.05$ are masked.}
    \label{fig:overlap_corr_heatmap}
\end{figure}

\paragraph{Effect of Longer Shared Tokens}
\Cref{fig:overlap_corr_heatmap} shows the Pearson correlation between token-overlap ratios of different lengths and downstream performance. The overlap ratio by length is computed as described in \Cref{eq:overlap_by_length}. 
As shown in \Cref{fig:overlap_corr_heatmap}, 
overlaps of relatively shorter tokens (including a single character) correlate negatively with performance, whereas overlaps of longer tokens show positive correlations. Notably, negative correlation between character-level overlaps and the performances indicates that while sharing characters alleviates the script barrier by reducing UNK tokens, excessive character-level overlap can be detrimental to downstream performance.
One possible explanation is that shorter tokens, which tend to vary more in meaning across contexts, may confuse the model when shared across languages, whereas longer tokens are more likely to provide more stable and consistent semantic cues across languages. 



\begin{figure}[t]
    \centering
    \includegraphics[width=\linewidth]{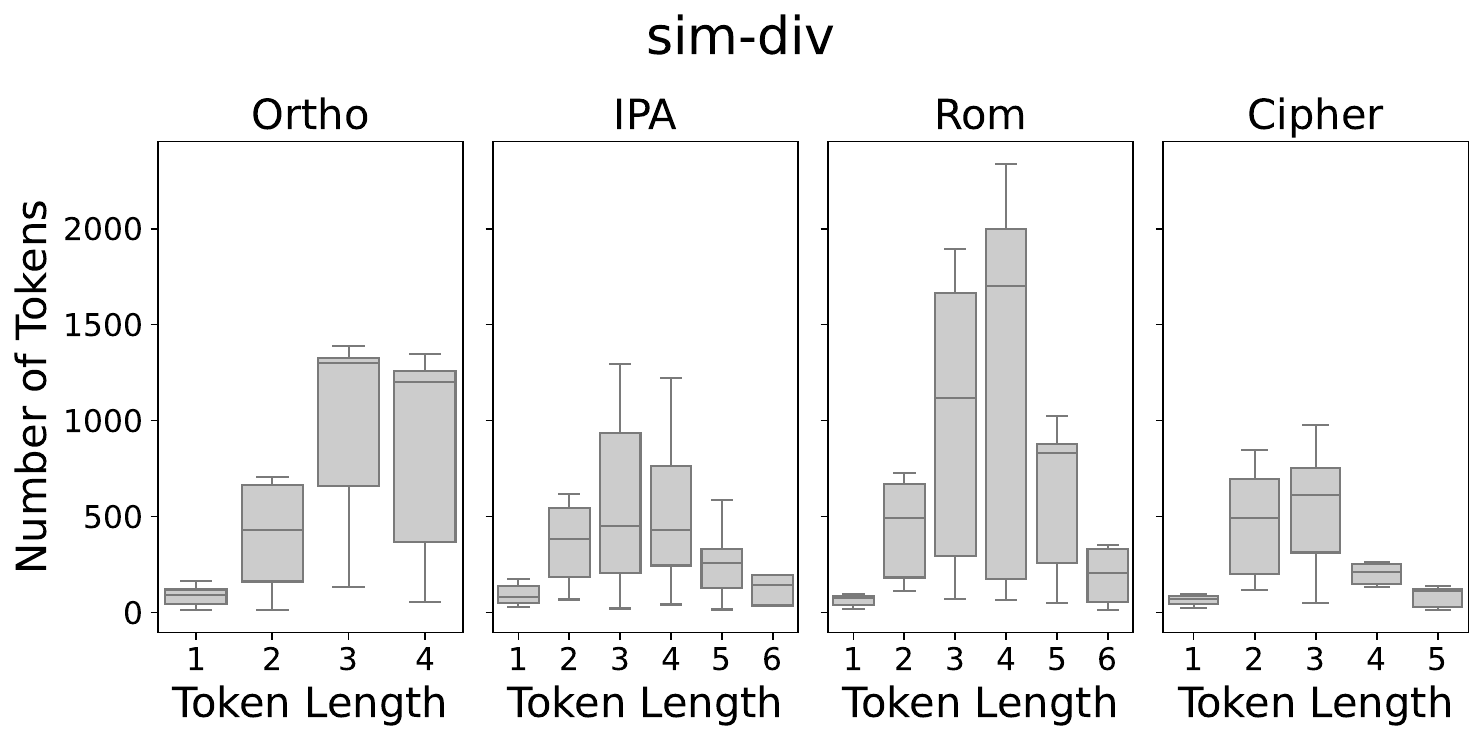}
    \caption{Number of unique tokens by length for unseen target languages, using models trained on sim-div languages.}
    \label{fig:wikiann_num_types_boxplot}
\end{figure}

\paragraph{Romanization and Token Length
}
We next examine how this observation relates to the notable success of romanization.
Our hypothesis is that romanization produces longer tokens, which in turn positively affect performance due to their better consistency across languages.
To test this hypothesis, we examine the distribution of unique token lengths produced by each model and find that Rom generates the largest proportion of longer tokens.
\Cref{fig:wikiann_num_types_boxplot} shows the distribution of token lengths for models trained on sim-div languages, with results for other language sets provided in the \Cref{appx:num_unique_tokens}.
We attribute this to the shared phonology introduced by transliteration, which we discuss further in \Cref{subsec:shared_phonology}.

\paragraph{Vocabulary Coverage}
To better understand the impact of longer shared tokens, we additionally analyze what we define as \textit{vocabulary coverage} and how each token length contribute to it. We define vocabulary coverage as the ratio of unique tokens produced to the total vocabulary size of the tokenizer. 
A higher coverage indicates that a larger portion of the model’s embedding space is being used, allowing more effective utilization of its capacity through a greater share of token embeddings.


\begin{figure}[]
    \centering
    \includegraphics[width=\linewidth]{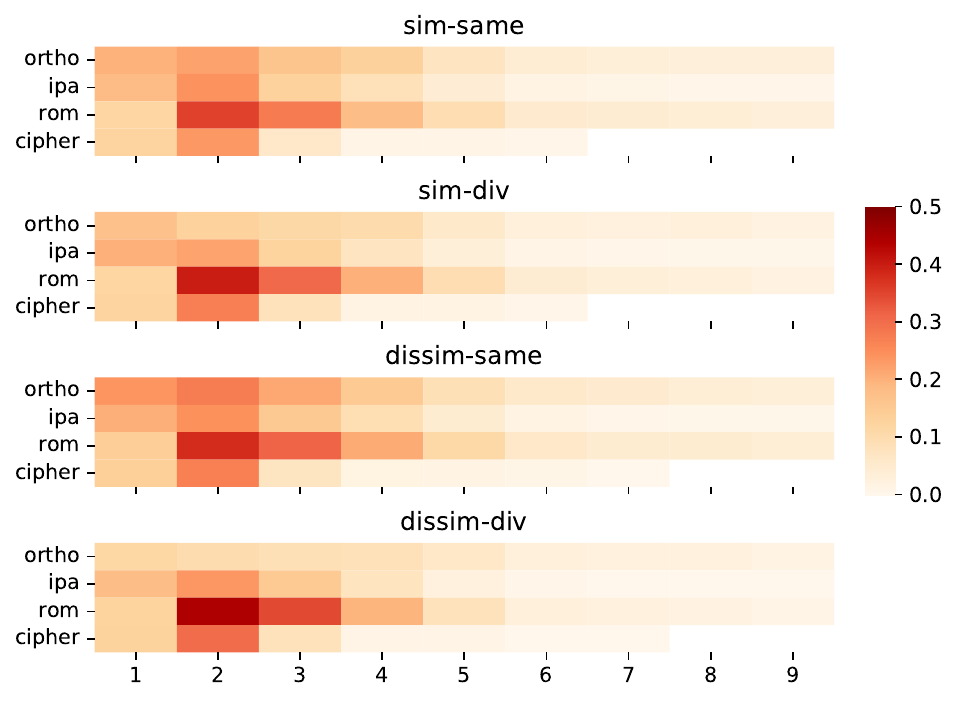}
    \caption{Average vocabulary coverage of each model on unseen languages. The x-axis indicates token length. Romanization shows higher coverage for tokens of length 2-4, whereas other input types exhibit more limited utilization of their vocabulary space. 
    }
    \label{fig:vocab_coverage_heatmap}
\end{figure}
\begin{figure}[t]
    \centering
    \includegraphics[width=\linewidth]{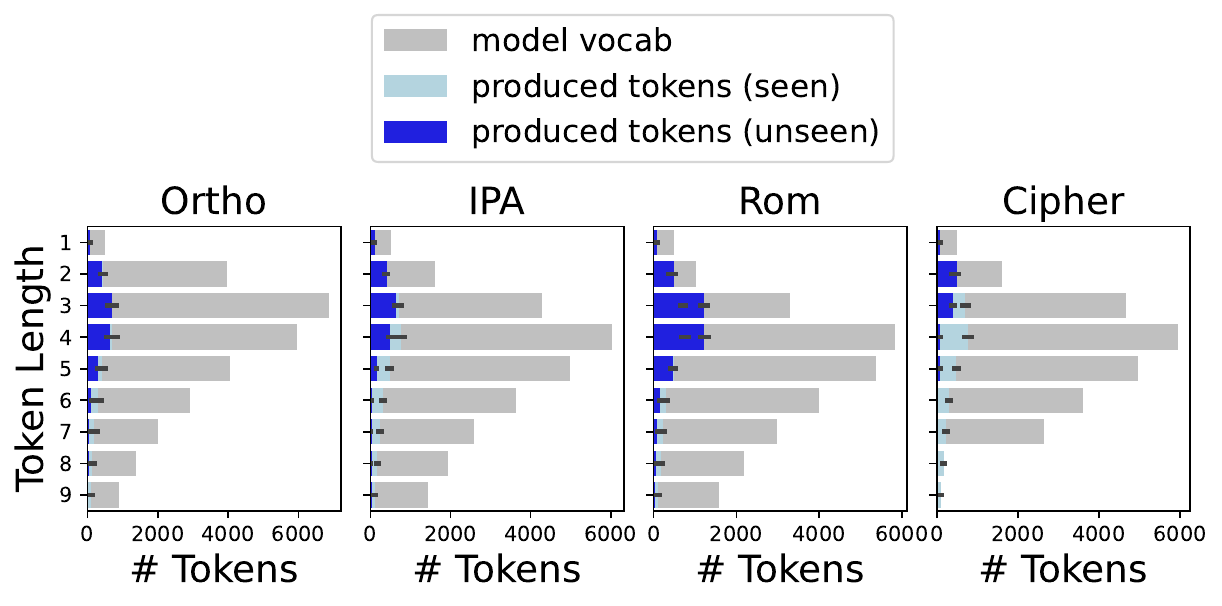}
    \caption{Number of tokens produced by each input type for seen and unseen languages. Rom utilizes a larger proportion of the model's vocabulary, whereas other input types show more limited token usage.}
    \label{fig:vocab_coverage}
\end{figure}

\Cref{fig:vocab_coverage_heatmap} shows the average vocabulary coverage across unseen languages. It clearly illustrates that Rom achieves higher coverage than other input types, especially for tokens longer than two characters.
\Cref{fig:vocab_coverage} further presents how the contribution of each token length to overall vocabulary coverage varies across input types. Rom clearly produces a greater variety of longer tokens than any other input type.
This broader token usage ensures greater vocabulary coverage and thereby better leverages the model's capacity.

\paragraph{Longer Tokens and Vocabulary Coverage} 
\begin{table}[h]
    \centering
    \resizebox{\columnwidth}{!}{\begin{tabular}{lr rr r}
    \toprule
    \multirow{2}{*}{\makecell{Input Type}}&\multicolumn{2}{c}{Fertility} & \multicolumn{2}{c}{Vocab. Coverage} \\ \cmidrule{2-5}
         & \multicolumn{1}{c}{Coef.} &\multicolumn{1}{c}{$p$}&\multicolumn{1}{c}{Coef.} &\multicolumn{1}{c}{$p$}\\\midrule
         Ortho & -0.2547 & 0.0008  & 0.5236 & $4.6 \times 10^{-12}$ \\
         IPA & -0.0176	& 0.8162 & 0.3248 & $1.8\times10^{-5}$\\
         Rom  &-0.0797 & 0.2927 & 0.3705 & $9.9 \times 10^{-7}$ \\
         Cipher & -0.1145 & 0.1303 &0.3405&$6.8\times 10^{-6}$\\
         All & -0.1628 &$1.2\times 10^{-5}$& 0.4240 & $5.0\times 10^{-30}$ \\\bottomrule
    \end{tabular}}
    \caption{Spearman’s correlation coefficients between downstream performance and two tokenizer-related factors: fertility and vocabulary coverage.}
    \label{tab:fertility_corr}
\end{table}

Building on the observation of how tokens of different lengths contribute to vocabulary coverage, we further interpret the advantage of overlapping longer tokens in terms of their combinatorial potential.
While single-character overlaps are bounded by the size of the alphabet or the number of unique characters available, longer tokens allow exponentially more possible combinations. This provides a greater potential to span a larger portion of the vocabulary, thereby enabling more effective utilization of the model’s capacity.

We also examine whether tokenizer fertility \citep{fertility, rust-etal-2021-good} explains the gains from transliteration. Fertility, defined as the average number of subword tokens a tokenizer generates per word, is commonly used to assess tokenizer quality. Prior work \citep{rust-etal-2021-good} suggests that lower fertility is associated with better multilingual performance.
Using our trained tokenizers, We compute fertility scores on the NER datasets and analyze their correlation with downstream performance. As shown in \Cref{tab:fertility_corr}, fertility only partially explains the performance gains and does not fully account for differences across input types. In contrast, vocabulary coverage exhibits a stronger and more consistent correlation---not only across input types, but also across languages within each input type. These results show that the benefit of sharing longer tokens for unseen languages cannot be explained solely by conventional tokenizer quality.

\subsection{Shared Phonology: A Path to Longer Tokens}\label{subsec:shared_phonology}
Here, we examine the role of shared phonology introduced by transliteration through the behavior of Cipher.
Cipher is designed to isolate the effect of shared phonology from Rom. It uses the same character set as in Rom, but does not encode phonological---or any linguistic---information shared across languages.

Without shared phonology, Cipher struggles to produce longer tokens as shown in \Cref{fig:wikiann_num_types_boxplot}, and consequently performs worse than Rom. Nevertheless, by eliminating unknown tokens, it performs comparably to IPA and surpasses Ortho on unseen languages.
In contrast, although IPA has a relatively high proportion of UNK tokens compared to other transliteration methods, it produces longer shared tokens in unseen languages. 
These observations suggest that the shared character set alone is insufficient for effective transliteration; rather, a consistent form-meaning mapping across languages is crucial to enable models to form longer tokens, typically through shared phonology.

\section{Conclusion}
Transliteration has been used to bring together diverse languages using diverse scripts, but its benefits have not been well understood.
We examined three aspects of transliteration---shared characters, shared subword tokens, and shared phonology---and analyzed their roles in adapting to languages and scripts unseen during pre-training, by comparing four different input types: Ortho, IPA, Rom, and Cipher.
We find that Rom outperforms all other input types, with a greater number of longer tokens shared across languages. 
While shared characters largely facilitate the adaptation by preventing unknown tokens, 
we suggest that sharing 
longer tokens is crucial for effective transliteration that overcomes the script barrier.
Our findings suggest that transliteration is effective not simply because of language similarity with pre-trained languages, but because it reshapes token distributions in ways that make multilingual models more adaptable and capable.
\section{Limitation}
The results reported here are suggestive, but there are two major limitations which prevent us from generalizing them too broadly. First, we only tested one type of transformer model with only subword tokenization scheme. It is possible, for example, that we would have obtained much different results if we had trained character- or byte-level models.
Also, we only tested one romanizer and one G2P transducer. It is possible that the results were influenced by the performance of each tool.

\section{Ethics Statement}

We believe that this research raises no significant ethical concerns or violations of the code of ethics mandated by the Association for Computational Linguistics. The data used in this study, all of which are publicly available, were collected in accordance with legal and institutional protocols, to the best of our knowledge. Furthermore, our use of these resources is compatible with the uses intended by the creators.
\section{Acknowledgement}
This research was supported by Korean Institute for Advancement of Technology(KIAT) grant funded by the Korea Government(MOTIE)(RS-2024-00435997, Human Resource Development Program for Industrial Innovation(Global)).
This work was also supported by the Institute of Information \& Communications Technology Planning\&Evaluation(IITP) grant funded by the Korea government(MSIT) (No. RS-2024-00437337, Development of technology for linking and utilizing security information and event management with privacy-preserving internal security data), as well as the University of British Columbia's Institute for Computing, Information and Cognitive Systems (ICICS).

\bibliography{anthology,custom}

@article{roberta,
  title={RoBERTa: A Robustly Optimized BERT Pretraining Approach},
  author={Yinhan Liu and Myle Ott and Naman Goyal and Jingfei Du and Mandar Joshi and Danqi Chen and Omer Levy and Mike Lewis and Luke Zettlemoyer and Veselin Stoyanov},
  journal={ArXiv},
  year={2019},
  volume={abs/1907.11692},
  url={https://api.semanticscholar.org/CorpusID:198953378}
}

@book{codebreakers@kahn,
  author = {David Kahn},
  title = {The Codebreakers: The Comprehensive History of Secret Communication from Ancient Times to the Internet},
  publisher = {Scribner},
  year = {1996},
  edition = {Revised},
  address = {New York}
}

@misc{fertility,
    title = {Exploring BERT's Vocabulary},
    author = {Judit Ács},
    note = {Blog Post},
    month = feb,
    year = {2019},
    urldate = {2019-02-19},
    url = {https://juditacs.github.io/2019/02/19/bert-tokenization-stats.html}
}

@inproceedings{nguyen-etal-2025-prompting,
    title = "Prompting with Phonemes: Enhancing {LLM}s' Multilinguality for Non-{L}atin Script Languages",
    author = "Nguyen, Hoang H  and
      Mahajan, Khyati  and
      Yadav, Vikas  and
      Salazar, Julian  and
      Yu, Philip S.  and
      Hashemi, Masoud  and
      Maheshwary, Rishabh",
    editor = "Chiruzzo, Luis  and
      Ritter, Alan  and
      Wang, Lu",
    booktitle = "Proceedings of the 2025 Conference of the Nations of the Americas Chapter of the Association for Computational Linguistics: Human Language Technologies (Volume 1: Long Papers)",
    month = apr,
    year = "2025",
    address = "Albuquerque, New Mexico",
    publisher = "Association for Computational Linguistics",
    url = "https://aclanthology.org/2025.naacl-long.599/",
    doi = "10.18653/v1/2025.naacl-long.599",
    pages = "11975--11994",
    ISBN = "979-8-89176-189-6"
}

@inproceedings{husain-etal-2024-romansetu,
    title = "{R}oman{S}etu: Efficiently unlocking multilingual capabilities of Large Language Models via {R}omanization",
    author = "Husain, Jaavid  and
      Dabre, Raj  and
      M, Aswanth  and
      Gala, Jay  and
      Jayakumar, Thanmay  and
      Puduppully, Ratish  and
      Kunchukuttan, Anoop",
    editor = "Ku, Lun-Wei  and
      Martins, Andre  and
      Srikumar, Vivek",
    booktitle = "Proceedings of the 62nd Annual Meeting of the Association for Computational Linguistics (Volume 1: Long Papers)",
    month = aug,
    year = "2024",
    address = "Bangkok, Thailand",
    publisher = "Association for Computational Linguistics",
    url = "https://aclanthology.org/2024.acl-long.833/",
    doi = "10.18653/v1/2024.acl-long.833",
    pages = "15593--15615"
}

\appendix

\section{Appendix}
\label{sec:appendix}

\subsection{Language Selection} 
\label{app:language_selection}
To examine the impact on multilingual adaptation that differences in input types have, we selected four language sets : (i) similar languages using the same script (sim-same), (ii) similar languages using diverse scripts (sim-div), (iii) dissimilar languages using the same script (dissim-same), and (iv) dissimilar languages using diverse scripts (dissim-div). These sets were used to train multilingual models with varying linguistic similarities and scripts. For each set, we assigned eight languages based on a computed similarity score as shown in \Cref{tab:language_sets}.

Similar to \citet{chang-etal-2024-multilinguality}, we utilized lang2vec \cite{littell-etal-2017-uriel}\footnote{Utilizing https://github.com/antonisa/lang2vec} to compute language similarity. Specifically, we extracted syntactic, geographic, and genetic features from lang2vec and computed cosine similarities, denoted as $\text{s}_{syn}$, $\text{s}_{geo}$, and $\text{s}_{gen}$ in Eq. \ref{eq:lang_similarity}. We also defined lexical similarity $\text{s}_{lex}$, which is obtained by calculating the word overlap ratio between training corpora of each language\footnote{Words are segmented by white spaces.}.
Finally, we aggregated all similarity scores (i.e., syntactic, geographic, genetic, and lexical) to derive the overall similarity score between two languages:

\begin{equation}
\label{eq:lang_similarity}
\begin{aligned}
\text{sim}_s(x,y) &= \text{s}_{syn}(x,y) 
                   + \text{s}_{geo}(x,y) \\
                  &\quad + \text{s}_{gen}(x,y)
                   + \text{s}_{lex}(x,y).
\end{aligned}
\end{equation}

With initial set of languages $L$ that are supported by Wikipedia corpus and Epitran, we use average pairwise similarity scores to compute similarity score for a set of languages and obtain an optimal set $L^*_s$, where $s \in \{\text{sim-same}, \text{sim-div}\}$ : 


\begin{equation}
\label{eq:langset_selection}
\scalebox{0.8}{$%
\begin{aligned}
L^*_s 
&= \arg\max_{\substack{L_s \subset L \\ |L_s|=8}} 
\biggl(
    \frac{1}{|L_s|(|L_s|-1)} 
    \sum_{x \in L_s} \sum_{\substack{y \in L_s \\ y \ne x}} \text{sim}_s(x, y)
\\
&\quad
+ \alpha \cdot 
  \Bigl(
    \mathbbm{1}_{s\in\{\text{sim-div}\}} |\text{SC}_{L_s}|
-
    \mathbbm{1}_{s\in\{\text{dissim-div}\}} |\text{SC}_{L_s}|
  \Bigr)
\biggr) \\
\end{aligned}
$}
\end{equation}.

As for an optimal set $L^*_d$, where $d \in \{\text{dissim-same}, \text{dissim-div}\}$ :

\begin{equation}
\label{eq:langset_selection_dissim}
\scalebox{0.8}{$%
\begin{aligned}
L^*_d 
&= \arg\min_{\substack{L_d \subset L \\ |L_d|=8}} 
\Biggl(
    \frac{1}{|L_d|(|L_d|-1)} 
    \sum_{x \in L_d} 
    \sum_{\substack{y \in L_d \\ y \ne x}} \text{sim}_s(x, y)
\\
&\quad
+ \alpha \cdot 
  \Bigl(
    \mathbbm{1}_{d\in\{\text{sim-div}\}} \,|\text{SC}_{L_d}|
-
    \mathbbm{1}_{d\in\{\text{dissim-div}\}} \,|\text{SC}_{L_d}|
  \Bigr)
\Biggr).
\end{aligned}
$}
\end{equation}
To select languages for the sets with same script (i.e., sim-same and dissim-same), we limited the search space to languages that use the Latin script to maximize the number of languages available for similarity-based sampling.

For sets with diverse scripts (i.e., -div), we additionally consider how many different scripts are involved in each set. 




\subsection{Model Configuration}
\label{app:model}

Table \ref{tab:model-config} summarizes the key configuration details of our RoBERTa-based model. Number of parameters per model is 109,082,112.

\begin{table}[ht]
\centering
\begin{tabular}{l l}
\hline
\textbf{Parameter}          & \textbf{Value} \\
\hline
Vocabulary Size             & 30,000 \\
Hidden Size                 & 768 \\
Hidden Layers               & 12 \\
Attention Heads             & 12 \\
Intermediate Size           & 3072 \\
Activation Function         & GELU \\
Dropout (Hidden/Attention)  & 0.1 \\
Max Position Embeddings     & 514 \\
\hline
\end{tabular}
\caption{Model Configuration}
\label{tab:model-config}
\end{table}

\subsection{Training Setup}
\label{app:training}
To investigate the impact of different input types, we pre-trained and fine-tuned a total of 16 models across four distinct input types and language sets. In addition, we trained a SentencePiece BPE tokenizer for each model, fixing the vocabulary size to 30K. Table \ref{tab:hyperparameters} summarizes the key hyperparameters used in our experiments for both the pretraining phase and the downstream NER task.

\paragraph{Hyperparameter Sweep}
We conducted grid search to find learning rates that converges or achieves the best results. For pre-training, the search space was \{1e-5, 2e-5, 3e-5, 5e-5, 1e-4, 2e-4, 3e-4\} and for NER, it was \{3e-5, 5e-5, 1e-4\}.

\begin{table}[ht]
\centering
\resizebox{\columnwidth}{!}{
\begin{tabular}{l c c c c}
\hline
\textbf{Parameter}                   & \textbf{Pretraining} & \textbf{NER} & \textbf{XNLI} &\textbf{POS}\\
\hline
FP16 Training                        & True               & True & True &True\\
Max Seq. Length                  & 512                & 512  & 512 & 512\\
\makecell[l]{Batch Size\\(per device)}              & 64                 & 64  & 64 & 128\\
\makecell[l]{Gradient \\Accumulation Steps}          & 1                  & -  & - & -\\
Warmup Steps                         & 50                 & - & - & -\\
Learning Rate                        & 1e-4               & 5e-5 & 3e-5 & 5e-5\\
Weight Decay                         & 0.01               & 0.01 & 0.01&0.01\\
LR Scheduler Type                    & Linear             & - & - & -\\
MLM Probability                      & 0.15               & - & - & -\\
Epochs                               & 300                & 20 & 5 & 20\\
Log Interval                         & -                & 1 & 1 & 1\\
\hline
GPU Resources & \makecell{4\\NVIDIA\\L40S} & \makecell{2\\NVIDIA\\RTX A6000} & \makecell{1\\NVIDIA\\RTX A6000} & \makecell{1\\NVIDIA\\RTX A6000}\\
\hline
\end{tabular}
}
\caption{Training Configurations}
\label{tab:hyperparameters}
\end{table}


\subsection{Substitution Cipher (Cipher)}
\label{app:cipher}
A substitution cipher is a method from cryptography where units of plaintext are replaced with ciphertext according to a predefined rule or key. We apply substitution cipher to the Romanized text to remove encoded phonological information. 

Specifically, we use the Caesar cipher \cite{codebreakers@kahn}, a simple substitution encryption technique that shifts each letter in the text by a fixed number of positions in the Latin alphabet. For each language, we assign an integer that determines the shift from the current position of each letter. For example, if English is assigned the integer 4, the word `apple' would be represented as `ettpi', with each letter replaced by the one four positions ahead in the alphabet.


\subsection{$P$-values of Paired t-tests}
\label{app:ttest-pvalue}

\begin{figure}
    \centering
    \begin{subfigure}[]{\linewidth}
        \includegraphics[width=\linewidth]{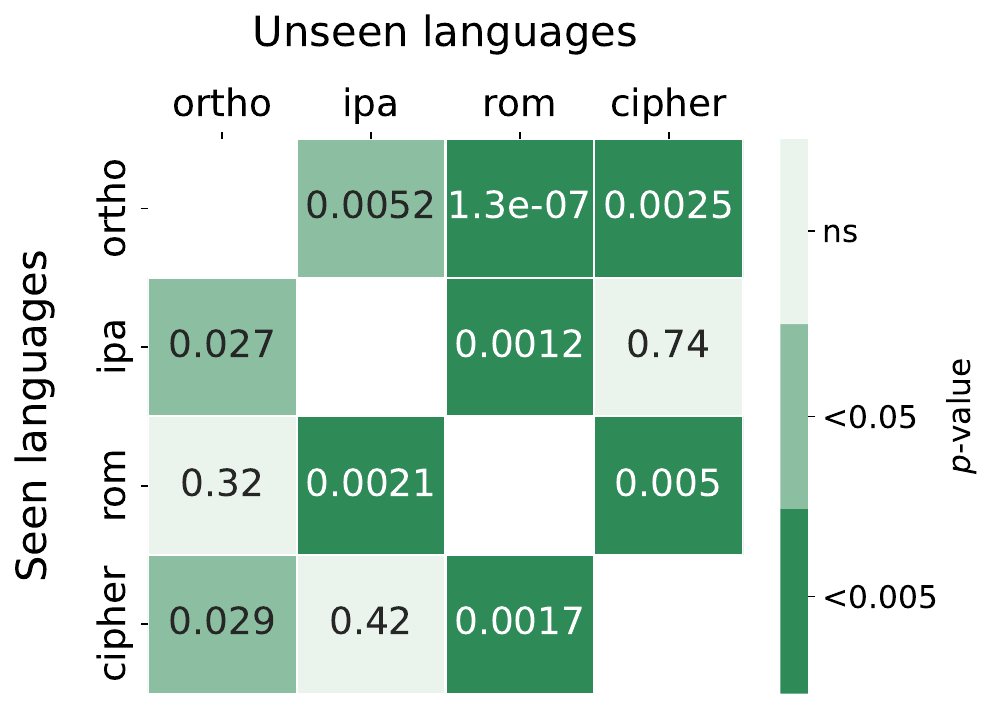}
        \caption{NER (WikiAnn)}
    \end{subfigure}
    \begin{subfigure}[]{\linewidth}
        \includegraphics[width=\linewidth]{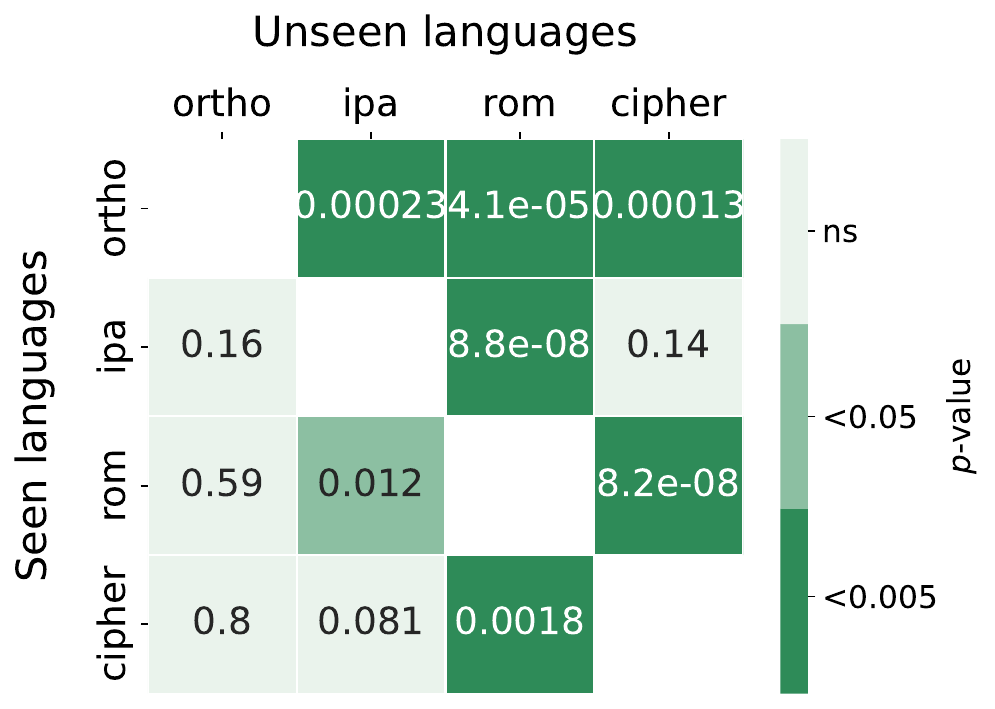}
        \caption{POS (Universal Dependencies)}
    \end{subfigure}
    \begin{subfigure}[]{\linewidth}
        \includegraphics[width=\linewidth]{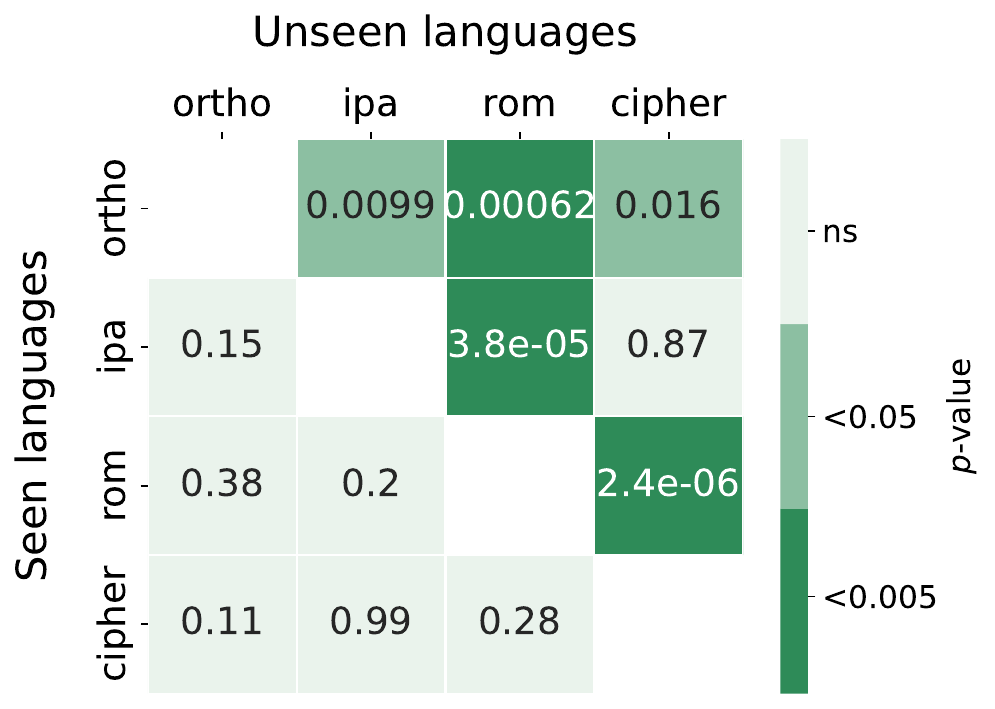}
        \caption{XNLI}
    \end{subfigure}
    \caption{$P$-values for paired $t$-tests on performances for all tasks. Upper triangular elements show results across unseen languages and lower triangular elements show those across seen languages.}
    \label{fig:ttest-pvalue}
\end{figure}

Table \ref{tab:ner} presents the scores of NER and XNLI tasks for different input types across various language settings. To assess the significance of the observed differences, we performed paired t-tests. 
To be specific, we run a paired, two-sided t-test, to allow for differences in either direction.
For each pair of input types (e.g., (Rom, Ortho), (Ortho, Cipher), etc.), we take the scores for all languages available in the task.
The number of observations for each test varies depending on the languages available for each task. For example, in NER, the t-test for unseen languages (p-values in the upper triangle in \Cref{fig:ttest-pvalue} (a)) uses 81 observations, and for seen languages (p-values in the lower triangle of the same figure), 31. We select observations in the same manner in other tasks as well.
Figure \ref{fig:ttest-pvalue} displays the corresponding $p$-values derived from these tests. The results show that transliteration (IPA, Rom, Cipher) significantly benefits Ortho for unseen languages. For NER, Rom and Ortho perform on par for seen languages, while there is no significant difference across any input types for XNLI.

\subsection{Multi-seed Results}
We conducted experiments with multiple random seeds for NER and POS to improve the reliability of the results. For each training configuration, we trained the models three times and averaged the F1 scores. The results show patterns consistent with our main observations (see \Cref{tab:multi-seed}).

\begin{table}[h]
    \centering
    \resizebox{\columnwidth}{!}{\begin{tabular}{llrrrrrr}
        \toprule
         \multirow{2}{*}{\makecell{Trained\\Lang. Set}} & \multirow{2}{*}{\makecell{Input\\Type}} & \multicolumn{3}{c}{NER} & \multicolumn{3}{c}{POS} \\\cmidrule{3-8}
         &&\multicolumn{1}{c}{All}&\multicolumn{1}{c}{Seen} & \multicolumn{1}{c}{Unseen}&\multicolumn{1}{c}{All} &\multicolumn{1}{c}{Seen} & \multicolumn{1}{c}{Unseen}  \\\midrule
         \multirow{5}{*}{sim-same}& \# Lang. & 28 & 8&20 & 19 & 7&12\\ \cmidrule{2-8}
         &Ortho &0.6954&0.8277&0.6425&0.8336&\textbf{0.9471}&0.7673 \\
         &IPA&0.7046&0.7984&0.6670&0.8936&0.9368&0.8684\\
         &Rom&\textbf{0.7438}&\textbf{0.8340}&\textbf{0.7077}&\textbf{0.9055}&0.9457&\textbf{0.8820}\\
         &Cipher&0.7084&0.8089&0.6682&0.9009&0.9450&0.8751\\\midrule
         \multirow{5}{*}{sim-div}&\# Lang.&28&8&20&19&5&14\\\cmidrule{2-8}
         &Ortho&0.6817&0.8281&0.6232&0.8107&\textbf{0.9499}&0.7610\\
         &IPA&0.7047&0.8131&0.6614&0.8913&0.9443&0.8724\\
         &Rom&\textbf{0.7437}&\textbf{0.8379}&\textbf{0.7061}&\textbf{0.8981}&0.9484&\textbf{0.8802}\\
         &Cipher&0.7041&0.8176&0.6587&0.8900&0.9462&0.8700\\\midrule
         \multirow{5}{*}{dissim-same}&\# Lang.&28&7&21&19&4&15\\\cmidrule{2-8}
         &Ortho&0.6895&\textbf{0.7902}&0.6559&0.8180&0.8946&0.7976\\
         &IPA&0.7109&0.7654&0.6928&0.8726&0.8666&0.8742\\
         &Rom&\textbf{0.7338}&0.7873&\textbf{0.7159}&\textbf{0.8911}&\textbf{0.8951}&\textbf{0.8901}\\
         &Cipher&0.7234&0.7614&0.7113&0.8836&0.8908&0.8817\\\midrule
         \multirow{5}{*}{dissim-div}&\# Lang.&28&8&20&19&5&14\\\cmidrule{2-8}
         &Ortho&0.7312&0.7239&0.7341&0.8406&0.8837&0.8252\\
         &IPA&0.7414&0.7394&0.7422&0.8996&\textbf{0.8921}&0.9023\\
         &Rom&\textbf{0.7627}&0.7339&\textbf{0.7742}&\textbf{0.9061}&0.8909&\textbf{0.9115}\\
         &Cipher&0.7406&\textbf{0.7403}&0.7407&0.8994&0.8905&0.9026\\\bottomrule
         
    \end{tabular}}
    \caption{Multi-seeded results for NER and POS. Results are averaged over three runs with different random seeds. \textbf{Bold} indicates best performing input type per language set.}
    \label{tab:multi-seed}
\end{table}

\subsection{Vocabulary Overlap Analysis} \label{appx:diff_compute}
\begin{figure}
    \centering
    \begin{subfigure}[]{\linewidth}
        \includegraphics[width=\linewidth]{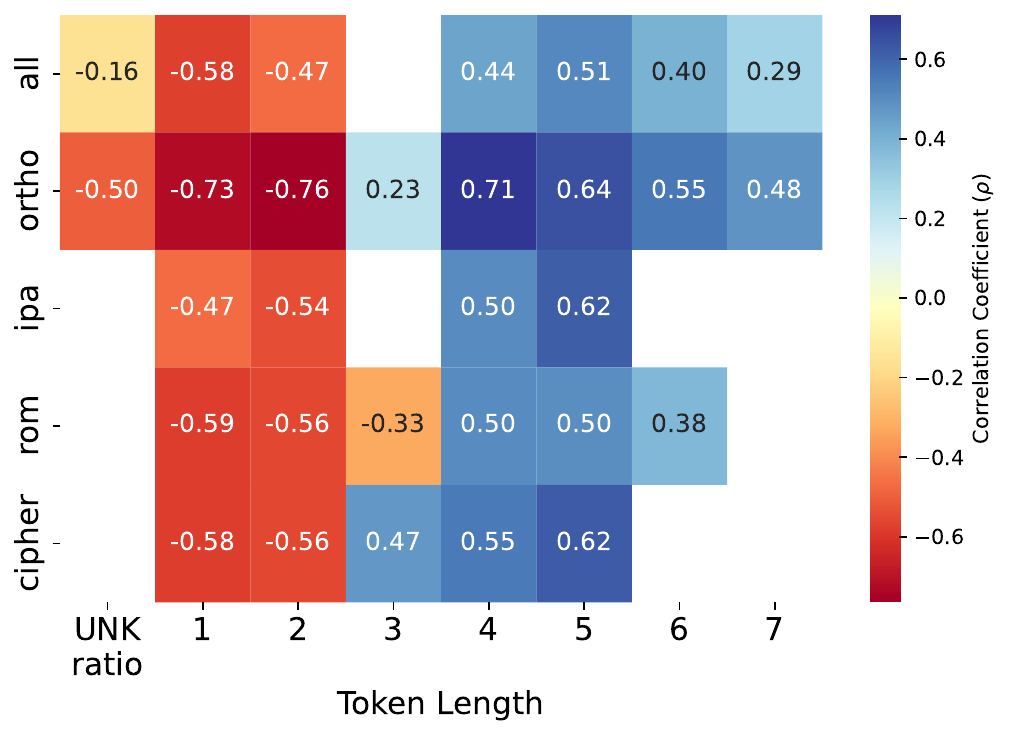}
        \caption{}
    \end{subfigure}
    \begin{subfigure}[]{\linewidth}
        \includegraphics[width=\linewidth]{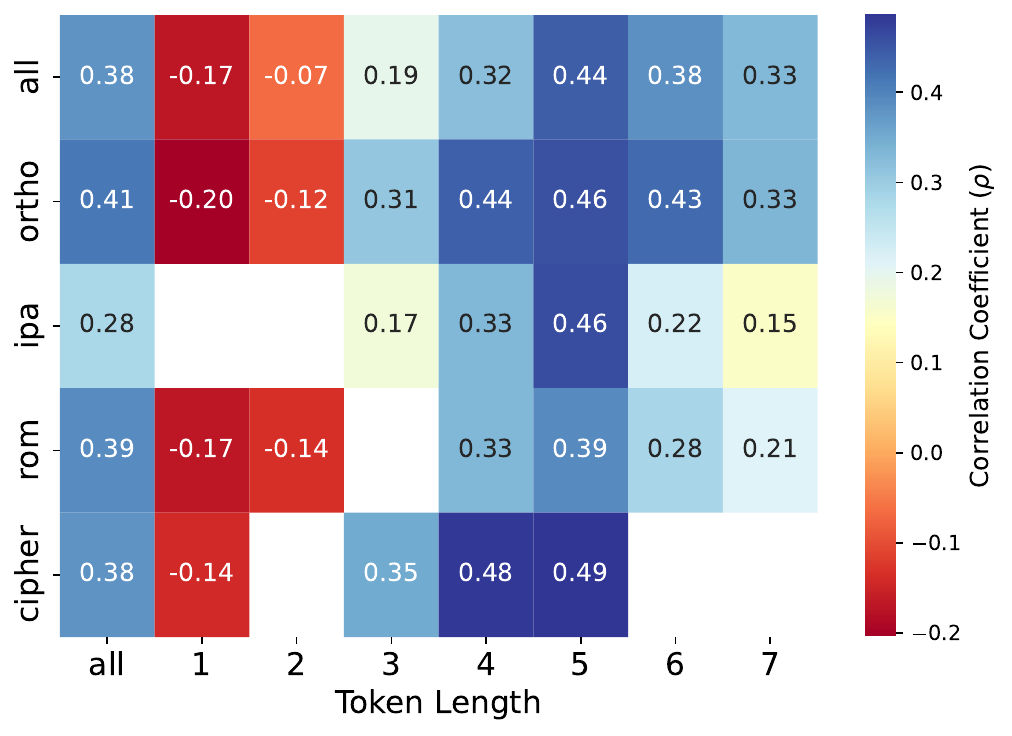}
        \caption{}
    \end{subfigure}
    
    \caption{Spearman's correlation coefficients between overlap ratio and NER F1 score. (a) Computed by type ratio. Leftmost column indicates correlations with UNK token ratio. (b) Computed by overlap ratio across all source languages. Leftmost column indicates correlations with total number of overlaps regardless of token length.}
    \label{fig:other_overlap_metric}
\end{figure}

\begin{figure}
    \centering
    \includegraphics[width=\linewidth]{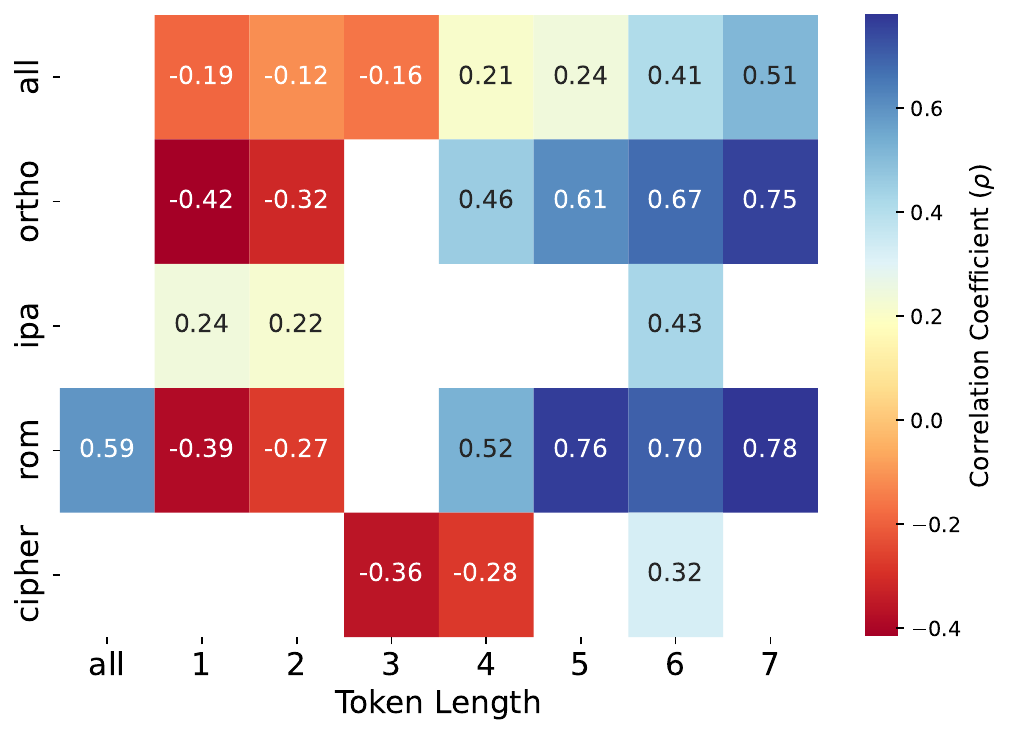}
    \caption{Spearman's correlation coefficients between overlap ratio and XNLI Accuracy. Overlap ratio computed as in \Cref{subsec:vocab_overlap}.}
    \label{fig:xnli_overlap_corr_heatmap}
\end{figure}

\begin{figure}
    \centering
    \includegraphics[width=\linewidth]{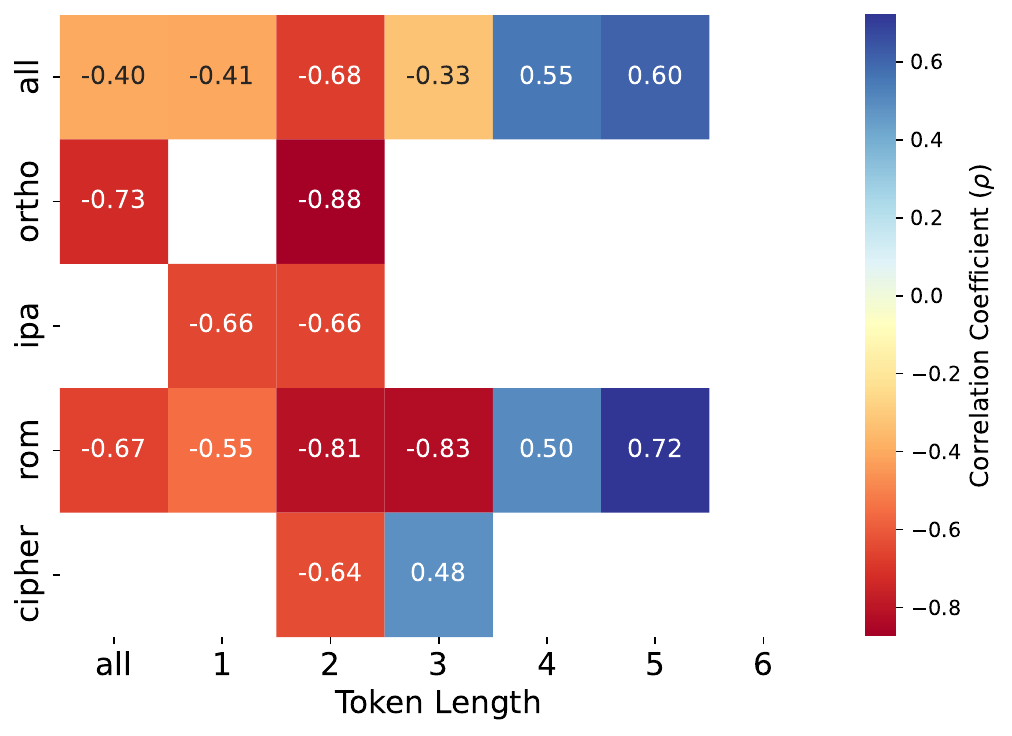}
    \caption{Spearman's correlation coefficients between overlap ratio and POS F1 score. Overlap ratio computed as in \Cref{subsec:vocab_overlap}.}
    \label{fig:pos_overlap_corr_heatmap}
\end{figure}

\Cref{fig:other_overlap_metric} shows heatmap of correlation coefficients between overlap ratio and NER F1 score, where overlap ratio is computed differently from \Cref{subsec:vocab_overlap}. \Cref{fig:xnli_overlap_corr_heatmap} shows the correlation analysis for XNLI task, and \Cref{fig:pos_overlap_corr_heatmap} for POS task.

\subsection{Number of Unique Tokens}\label{appx:num_unique_tokens}
\begin{figure}
    \centering
    \begin{subfigure}[]{\linewidth}
        \includegraphics[width=\linewidth]{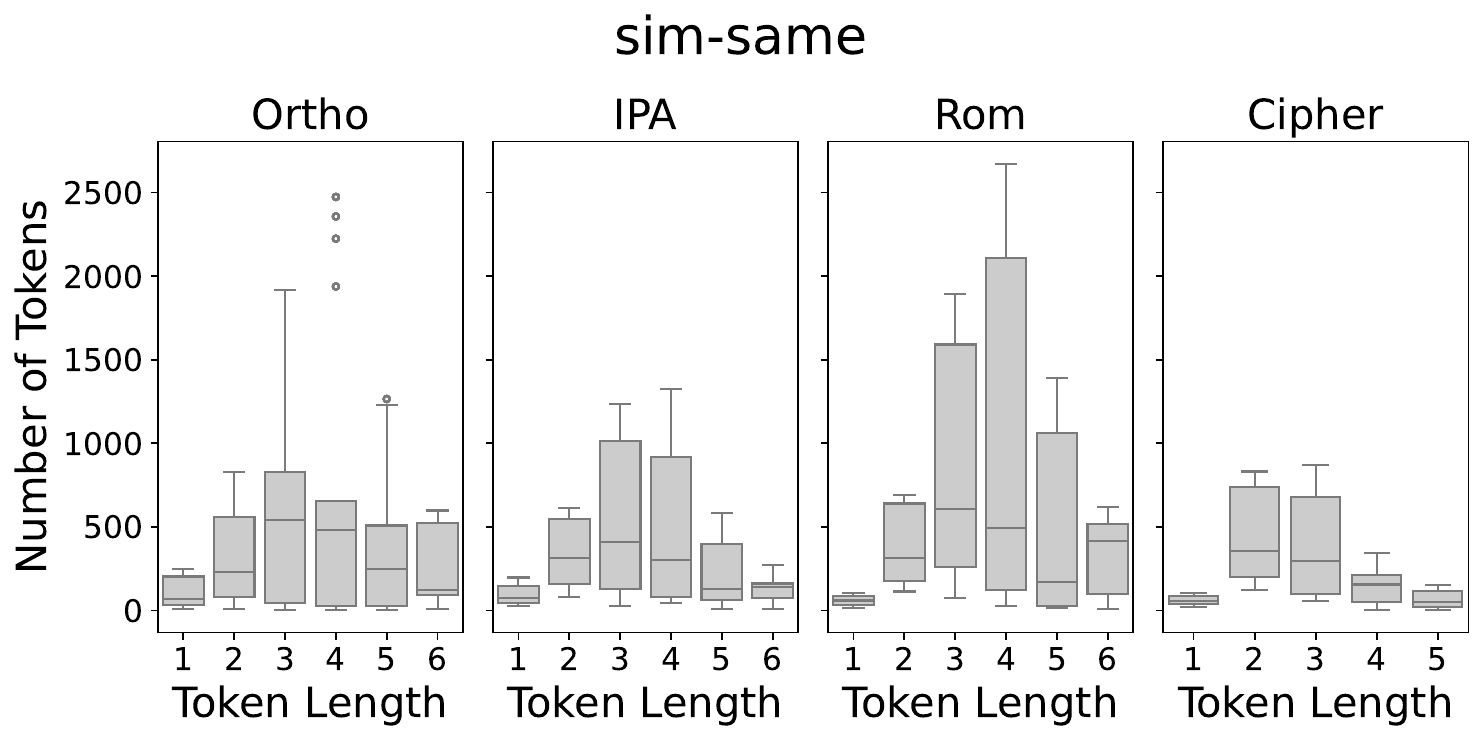}
    \caption{}    
    \end{subfigure}
    \begin{subfigure}[]{\linewidth}
        \includegraphics[width=\linewidth]{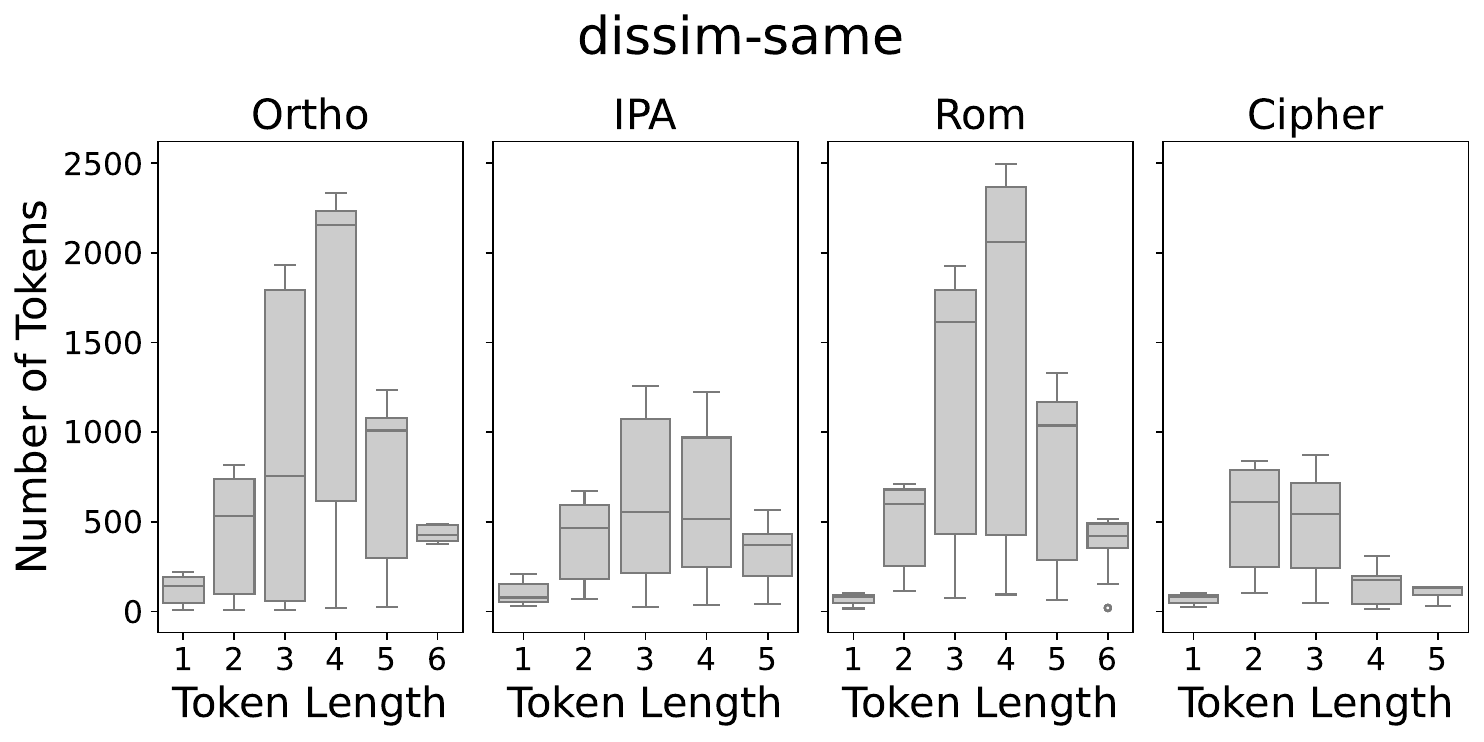}
    \caption{}    
    \end{subfigure}
    \begin{subfigure}[]{\linewidth}
        \includegraphics[width=\linewidth]{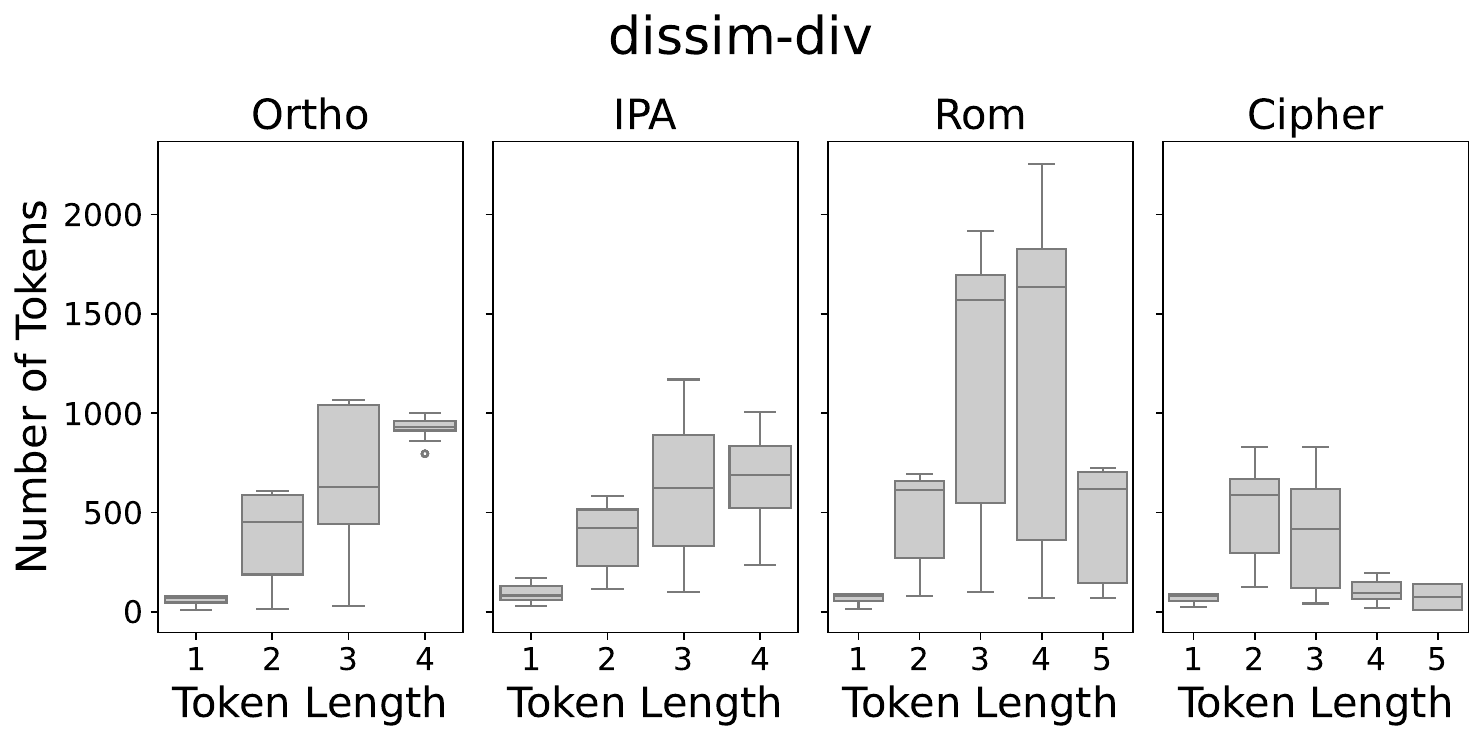}
    \caption{}    
    \end{subfigure}
    \caption{Number of unique tokens by length for unseen target languages (WikiAnn dataset), using models trained on (a) sim-same (b) dissim-same (c) dissim-div languages.}
    \label{fig:num_unique_tokens}
\end{figure}

\Cref{fig:num_unique_tokens} shows the distribution of token lengths for models trained on each language set.

\subsection{Vocabulary Coverage}
\begin{figure}
    \centering
    \includegraphics[width=\linewidth]{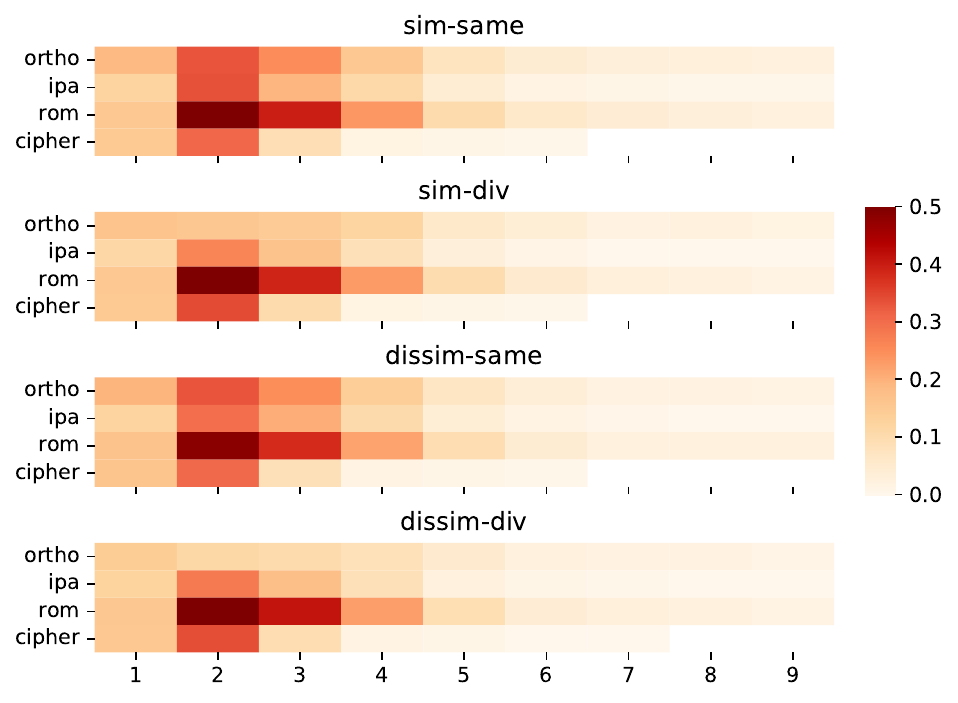}
    \caption{Vocabulary coverage over unseen languages on XNLI dataset computed as in \Cref{subsec:shared_token_set}.}
    \label{fig:xnli_vocab_coverage_heatmap}
\end{figure}

\begin{figure}
    \centering
    \includegraphics[width=\linewidth]{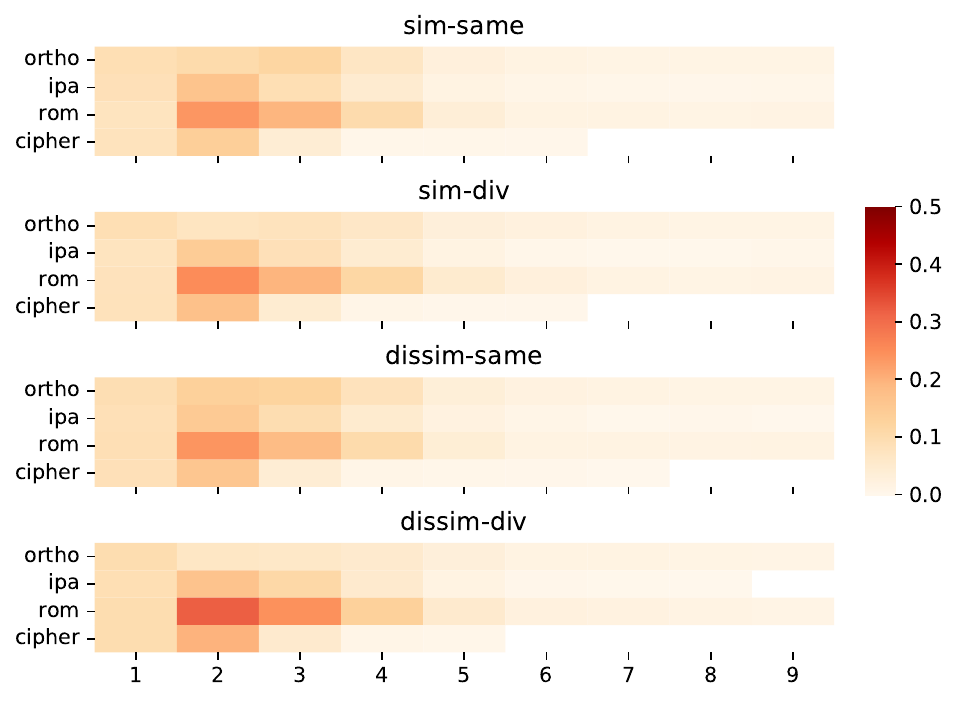}
    \caption{Vocabulary coverage over unseen languages on UD Treebank computed as in \Cref{subsec:shared_token_set}.}
    \label{fig:pos_vocab_coverage_heatmap}
\end{figure}

\begin{figure}
    \centering
    \includegraphics[width=\linewidth]{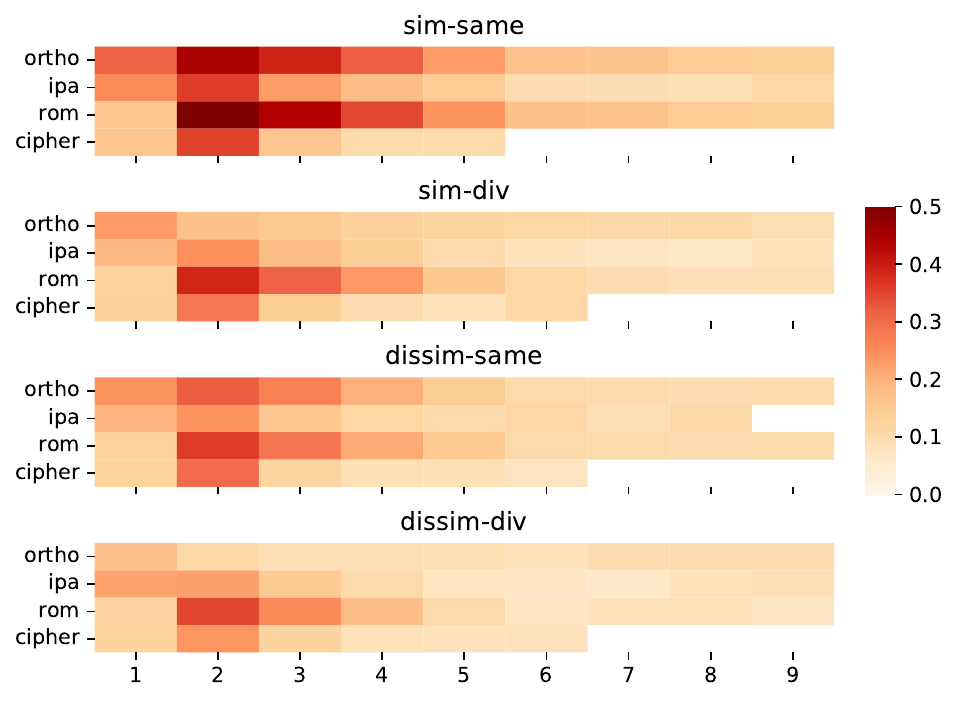}
    \caption{Vocabulary coverage over seen languages on WikiAnn dataset computed as in \Cref{subsec:shared_token_set}.}
    \label{fig:wikiann_vocab_coverage_heatmap}
\end{figure}

Here we provide more plots for vocabulary coverage. \Cref{fig:xnli_vocab_coverage_heatmap,fig:pos_vocab_coverage_heatmap} shows vocabulary coverage by length over unseen languages using XNLI dataset and UD Treebank, respectively. The pattern is similar to that of NER (WikiAnn) dataset, where Rom exhibits evidently high coverage compared to other input types. \Cref{fig:wikiann_vocab_coverage_heatmap} shows vocabulary coverage by length over seen languages using WikiAnn dataset. While Rom still shows the most coverage, Ortho also takes large portion when the languages share same script (i.e., sim-same, dissim-same).


\subsection{Tokenization Algorithm: Unigram}
\begin{table}[]
    \centering
    \resizebox{\columnwidth}{!}{
    \begin{tabular}{llrrr}
    \toprule
         \makecell[c]{Trained \\Lang. Set}&\makecell[c]{Input Type}&\makecell[c]{All}&\makecell[c]{Seen}&\makecell[c]{Unseen}   \\\midrule
         \multirow{4}{*}{sim-same} & Ortho & 0.5472& 0.7717& 0.4575 \\
         &IPA&0.5777&0.7024&0.5278 \\
         &Rom&0.7092&0.7908&0.6766 \\
         &Cipher & 0.6162 &0.6694&0.5949 \\\midrule
         \multirow{4}{*}{sim-div}&Ortho&	0.6781&0.7894&0.6336 \\
         &IPA&0.5765&0.6708& 0.5387\\
         &Rom&0.7042&0.8204&0.6578 \\
         &Cipher&0.7433&0.8489&0.7011\\\bottomrule
    \end{tabular}
    }
    \caption{Average F1 scores for NER task using Unigram.}
    \label{tab:unigram}
\end{table}

\begin{table}[]
    \centering
    \resizebox{\columnwidth}{!}{
    \begin{tabular}{llrrrrr}
    \toprule
         \makecell[c]{Trained\\Lang. Set}&\makecell[c]{Input Type}&1&2&3&4&5  \\\midrule
         \multirow{2}{*}{sim-same}&IPA&0.0302&0.2635&0.1636&0.1014&0.0570\\
         &Cipher&0.1491&0.3450&0.0714&0.0147&0.0202\\\midrule
         \multirow{2}{*}{sim-div}&IPA&0.0304&0.1917&0.1592&0.0871&0.0483 \\
         &Cipher&0.1415&0.3543&0.0907&0.0185&\\\bottomrule
    \end{tabular}
    }
    \caption{Vocabulary coverage of Unigram tokenizer.
    }
    \label{tab:unigram_vocab_coverage}
\end{table}

We conducted additional pre-training experiments using the Unigram tokenizer. We focused on two language sets, sim-same and sim-div, to limit computational cost while preserving diversity in script configurations.

We ran two-sided paired $t$-test, as we did in the main analyses, to examine the statistical significance of the results. To summarize the statistically significant findings ($p<0.05$):
\begin{itemize}[nosep]
    \item Models trained on sim-same languages set, 
    \begin{itemize}[nosep]
        \item Rom was significantly better than IPA ($p<0.005$)
        \item Rom was significantly better than Ortho ($p<0.001$)
    \end{itemize}
    \item Models trained on sim-div language set,
    \begin{itemize}
        \item Rom was significantly better than IPA ($p<0.005$)
        \item Cipher was significantly better than IPA ($p<0.0005$)
    \end{itemize}
\end{itemize}

We note that Cipher behaves differently under the Unigram tokenizer compared to BPE, in that it often outperforms IPA. To investigate this, we analyzed the vocabulary coverage by token length as in \Cref{fig:vocab_coverage_heatmap}, and found that Cipher exhibits notably higher coverage than IPA, consuming more tokens for unseen languages (See \Cref{tab:unigram_vocab_coverage}).

We believe this difference arises from how Unigram constructs its vocabulary: rather than relying on greedy merges, it chooses subwords based on probability, resulting in a different token-length distribution in the vocabulary. While our main findings based on token length distributions still hold, this ablation illustrates that different tokenization algorithms may emphasize different aspects of each factor we defined.


\subsection{External Tools for Transliteration}
In this study, we used Epitran and Uroman as transliteration tools to unify script and facilitate multilingual processing. These tools are widely used for converting text into standardized phonemic or Romanized forms, which aids in cross-lingual learning and transferability. Below, we describe their functionalities and implementation details.

Epitran\cite{mortensen-etal-2018-epitran} is a tool for grapheme-to-phoneme (G2P) conversion, capable of converting text into the International Phonetic Alphabet (IPA) representations. It can be downloaded from the link below
\underline{https://github.com/dmort27/epitran}

Uroman\cite{hermjakob-etal-2018-box} is a universal transliteration tool that converts text from various scripts into a Romanized format. It can be downloaded from the link below \underline{https://github.com/isi-nlp/uroman}

\subsection{Datasets} 

\label{app:dataset}
In Table \ref{tab:dataset-pretrain}, the specific number of datasets per corresponding language is provided. For pre-training, we utilized sampled version of preprocessed Wikipedia corpus from Huggingface\footnote{https://huggingface.co/datasets/wikimedia/wikipedia}.

We limited each language with its number of words around 10M\footnote{For each language, we randomly shuffled the order of the documents, and iterated over each document, counting the words segmented by whitespaces. We stop adding the documents when adding the number of words of the last document exceeds 10M.}. For those languages with less number of tokens than 10M, we kept all the documents and oversampled during training, to match the model's exposure to all languages.
For downstream task, we utilized WikiAnn \cite{pan-etal-2017-cross, rahimi-etal-2019-massively} dataset for named entity recognition, Universal Dependencies Treekbank \cite{nivre-etal-2020-universal} for POS, and XNLI dataset for natural language inference (NLI) task. In order to train the model with different input types, we converted all datasets into each corresponding input type.

Wikipedia corpora used for pre-training are licensed under the GNU Free Documentation License (GFDL) and the Creative Commons Attribution-Share-Alike 3.0 License. License type for WikiAnn dataset is ODC-BY.

\begin{table*}[h]
    \centering
    \resizebox{\textwidth}{!}{%
    \begin{tabular}{l|c|ccc|l|c|ccc}
    \toprule
         Lang & Dataset & \# Train & \# Validate & \# Test & Lang & Dataset & \# Train & \# Validate & \# Test \\
         \midrule 
         \multirow{3}{*}{am} & WikiAnn & 100 & 100 & 100 & \multirow{3}{*}{my} & WikiAnn &  100 & 100 & 100\\
         & XNLI &- & -&- && XNLI &-&-&- \\
         &UD Treebank&-&-&-&&UD Treebank&-&-&-\\
         \midrule
         \multirow{3}{*}{bn} & WikiAnn &10000 & 1000 & 1000 & \multirow{3}{*}{or} & WikiAnn & 100 & 100 & 100\\
         & XNLI & - & - & - && XNLI &-&-&- \\
         &UD Treebank&-&-&-&&UD Treebank&-&-&-\\
         \midrule
         \multirow{3}{*}{ca} & WikiAnn & 20000 & 10000 & 10000& \multirow{3}{*}{pt}& WikiAnn & 20000 & 10000 & 10000\\
         & XNLI & - & - & - && XNLI &-&-&- \\
         &UD Treebank&13123&1709&1846&&UD Treebank&9616&1204&1200\\
         \midrule
         \multirow{3}{*}{de} & WikiAnn & 20000 & 10000 & 10000 & \multirow{3}{*}{ro} & WikiAnn &20000& 10000& 10000\\
         & XNLI & 392702 & 2490 & 5010 && XNLI &-&-&- \\
         &UD Treebank&13813&799&977&&UD Treebank&8043&752&729\\
         \midrule
         \multirow{3}{*}{es} & WikiAnn & 20000 & 10000 & 10000 &\multirow{3}{*}{ru} & WikiAnn & 20000& 10000& 10000\\
         & XNLI & 392702 & 2490 & 5010 && XNLI &392702 & 2490 & 5010\\
         &UD Treebank&14187&1400&427&&UD Treebank&3850&579&601\\
         \midrule
         \multirow{3}{*}{fi} & WikiAnn &20000 & 10000 & 10000 & \multirow{3}{*}{so}& WikiAnn & 100&100 &100 \\
         & XNLI & - & - & - && XNLI &-&-&- \\
         &UD Treebank&12217&1364&1555&&UD Treebank&-&-&-\\
         \midrule
         \multirow{3}{*}{fr} & WikiAnn & 20000 & 10000 & 10000& \multirow{3}{*}{sq}& WikiAnn &5000& 1000& 1000\\
         & XNLI & 392702 & 2490 & 5010 && XNLI &-&-&- \\
         &UD Treebank&14450&1476&416&&UD Treebank&160&20&20\\
         \midrule
        \multirow{3}{*}{hi} & WikiAnn &5000 & 1000 & 1000 & \multirow{3}{*}{sr}& WikiAnn &  20000 & 10000 & 10000\\
        & XNLI & 392702 & 2490 & 5010 && XNLI &-&-&- \\
         &UD Treebank&13306&1659&1684&&UD Treebank&-&-&-\\
         \midrule
         \multirow{3}{*}{hr}& WikiAnn & 20000 & 10000 & 10000& \multirow{3}{*}{sv}& WikiAnn &20000& 10000& 10000\\
         & XNLI & - & - & - && XNLI &-&-&- \\
         &UD Treebank&6914&960&1136&&UD Treebank&3457&1118&1121\\
         \midrule
         \multirow{3}{*}{ilo}& WikiAnn & 100 & 100 & 100& \multirow{3}{*}{sw}& WikiAnn & 1000 & 1000& 1000\\
         & XNLI & - & - & - && XNLI & 392702 & 2490 & 5010 \\
         &UD Treebank&-&-&-&&UD Treebank&-&-&-\\
         \midrule
         \multirow{3}{*}{ka}& WikiAnn & 10000 & 10000 & 10000 &\multirow{3}{*}{te} & WikiAnn &1000& 1000& 1000\\
         & XNLI & - & - & - && XNLI &-&-&- \\
         &UD Treebank&870&130&818&&UD Treebank&1051&131&146\\
         \midrule
         \multirow{3}{*}{ko}& WikiAnn &  20000 & 10000 & 10000&\multirow{3}{*}{th} & WikiAnn &20000& 10000& 10000\\
         & XNLI & - & - & - && XNLI & 392702 & 2490 & 5010 \\
         &UD Treebank&4400&950&989&&UD Treebank&2902&362&363\\
         \midrule
         \multirow{3}{*}{lij} & WikiAnn & 100 & 100 & 100 &\multirow{3}{*}{ur} & WikiAnn &20000 & 1000& 1000\\
         & XNLI & - & - & - && XNLI & 392702 & 2490 & 5010 \\
         &UD Treebank&-&-&-&&UD Treebank&4043&552&535\\
         \midrule
         \multirow{3}{*}{lv} & WikiAnn & 10000 & 10000 & 10000&\multirow{3}{*}{uz} & WikiAnn & 1000& 1000& 1000\\
         & XNLI & - & - & - && XNLI &-&-&- \\
         &UD Treebank&15058&2110&2412&&UD Treebank&483&-&198\\
         \bottomrule
    \end{tabular}
    }
    \caption{Statistics of the transliterated datasets for downstream tasks. All datasets exist in four parallel versions: original orthographic, phonemic IPA, Romanized, and cipher-transcribed. A dash (-) indicates unavailable values. Some languages, such as English, Turkish, and Vietnamese, are used only in the XNLI dataset, and the number of items is equal across all XNLI datasets.}

    \label{tab:dataset-pretrain}
\end{table*}


\subsection{Detailed Experimental Results}

Tables \ref{tab:appx_simsame}, \ref{tab:appx_simdiff}, \ref{tab:appx_dissimsame}, \ref{tab:appx_dissimdiff} summarize the performance results (F1 scores) across different language sets under various evaluation settings. In our experiments, "Seen" refers to languages included in both pretraining and fine-tuning, "Unseen" to those entirely absent during training, and "Zero-Shot" to languages evaluated without task-specific fine-tuning. The language sets differ in terms of typological similarity and script usage. Detailed results for each setting are provided in the respective tables.
\begin{table}[]
    \resizebox{\columnwidth}{!}{%
    \begin{tabular}{l|l|llll}
    \toprule
     & & Ortho    & IPA   & Rom   & Cipher  \\
     \midrule
     
     \multirow{8}{*}{Seen} & ca& 0.8997 & 0.8725& 0.8993 & 0.8803\\
     &es& 0.8773  & 0.8584& 0.8788 &0.8657\\
     &fr& 0.8628  &0.8252& 0.8639& 0.8384\\
     &lij& 0.5064  & 0.4052 & 0.4615& 0.4082 \\
     &pt& 0.8798 & 0.8605& 0.8796& 0.8674 \\
     &ro& 0.9129 & 0.8855& 0.9103& 0.8956\\
     &sq& 0.9120 & 0.8738& 0.8979& 0.8785 \\
     &sv& 0.9215 & 0.8872 & 0.9247 & 0.9046\\
     \midrule
     \multirow{21}{*}{Unseen} & am&0.2000 & 0.3089 & 0.3383 & 0.3623\\
     &bn&0.8230 &0.8907 & 0.9081 & 0.8969 \\
     &de& 0.8204& 0.7400 &  0.8236  & 0.7676 \\
     &fi&0.8573& 0.8050 & 0.8609 & 0.8237 \\
     &hi&0.7395 & 0.8043 & 0.8225 & 0.7861\\
     &hr&0.8682 & 0.8318 & 0.8727 & 0.8403\\
     &ilo& 0.6400 & 0.5714 & 0.6757 & 0.4498 \\
     & ka &0.6878 & 0.7920 & 0.8227 & 0.7780 \\
     & ko& 0.5329 & 0.7578 & 0.7883 & 0.7626 \\
     & lv& 0.8940 & 0.8463 & 0.8919 & 0.8695 \\
     & my& 0.2286& 0.2541 & 0.2857 & 0.2232 \\
     & or& 0.2738& 0.2647 & 0.3492 & 0.3533\\
     & ru&0.8083 & 0.7842 & 0.8268  & 0.8010 \\
     & sn& -& - & -& -\\
     & so& 0.6256 & 0.4641 & 0.5500 & 0.4397 \\
     & sr& 0.8574 & 0.8442 & 0.8879 & 0.8691 \\
     & sw& 0.8250 & 0.7381 & 0.8195 & 0.7494 \\
     & te&0.3384 & 0.5336 & 0.5797 & 0.5252 \\
     & th& 0.4762 & 0.6637 &0.6622 & 0.6477 \\
     & ur& 0.9032 & 0.9101 & 0.9273& 0.9172 \\
     & uz& 0.8266 & 0.7962 & 0.8402 & 0.7862 \\
     \bottomrule
    \end{tabular}%
    }
    \caption{F1 scores of each model trained on \textbf{sim-same} language set. \textbf{Seen}: languages used during both pre-training. \textbf{Unseen}: languages not encountered during pre-training. Results are provided for four different input types: Orthographic (Ortho), IPA, Romanized (Rom), and Cipher (Cipher).}
    \label{tab:appx_simsame}
\end{table}

\begin{table}[]
    \resizebox{\columnwidth}{!}{%
    \begin{tabular}{l|l|llll}
    \toprule
     & & Ortho    & IPA   & Rom   & Cipher  \\
     \midrule
     
     \multirow{8}{*}{Seen} & bn& 0.9380 & 0.9375 & 0.9466 & 0.9377 \\
     &fr & 0.8436 & 0.8255 & 0.8430 & 0.8378 \\
     & hi& 0.8524 & 0.8577 & 0.8394 & 0.8314\\
     & hr& 0.8741 & 0.8527 & 0.8767 & 0.8605 \\
     & or&0.5483 & 0.4981 & 0.5873 & 0.4962 \\
     & ru& 0.8395 & 0.8286 & 0.8375 & 0.8304 \\
     & sr& 0.8918 & 0.8484 & 0.8969 & 0.8900 \\
     & ur& 0.9396 & 0.9424 & 0.9333 & 0.9318 \\
     \midrule
     \multirow{21}{*}{Unseen} & am& 0.0079 & 0.2902 & 0.3282& 0.2695 \\
     &ca & 0.8769 &0.8399&0.8801&0.8443 \\
     & de& 0.7934 & 0.7381 & 0.8047 & 0.7608 \\
     &es& 0.8511 & 0.8144 & 0.8573 & 0.8265 \\
     & fi& 0.8427 & 0.7993 & 0.8460 & 0.8201 \\
     &ilo& 0.5333 & 0.5356 & 0.5537 & 0.4627 \\
     & ka&0.5860 & 0.7961 & 0.8162 & 0.7872 \\
     & ko& 0.5244 & 0.7318 & 0.7792 & 0.7577 \\
     &lij& 0.3071 & 0.3684 & 0.2975 & 0.3064 \\
     & lv& 0.8826 & 0.8468 & 0.8891 & 0.8605 \\
     & my& 0.1596 & 0.1721 & 0.2975 & 0.2424 \\
     & pt&0.8535 & 0.8206 & 0.8547 &  0.8312 \\
     & ro& 0.8889 & 0.8754 & 0.8963 &  0.8695 \\
     & sn&-&-&-&-\\
     & so& 0.4874 & 0.4870 & 0.5128 & 0.5236 \\
     & sq&  0.8557 & 0.8319 & 0.8604 &  0.8315 \\
     & sv&  0.9059 & 0.8583 & 0.9043 &  0.8850 \\
     & sw& 0.7634 & 0.7429 & 0.7955 & 0.7359 \\
     & te& 0.3297 & 0.5753 & 0.6440 & 0.5119 \\
     & th& 0.3531 & 0.6680& 0.6479 & 0.6302 \\
     & uz& 0.8384 & 0.7819& 0.8360 & 0.7863 \\
     \bottomrule
    \end{tabular}%
    }
    \caption{F1 scores of each model trained on \textbf{sim-div} language set. \textbf{Seen}: languages used during both pre-training. \textbf{Unseen}: languages not encountered during pre-training. Results are provided for four different input types: Orthographic (Ortho), IPA, Romanized (Rom), and Cipher (Cipher).}
    \label{tab:appx_simdiff}
\end{table}
\begin{table}[]
\resizebox{\columnwidth}{!}{%
    \begin{tabular}{l|l|llll}
    \toprule
     & & Ortho    & IPA   & Rom   & Cipher \\
     \midrule
     \multirow{8}{*}{Seen} &de& 0.8184 & 0.7924 & 0.8248 & 0.8095\\
     &fi&0.8618 & 0.8264 & 0.8638 & 0.8436\\
     &ilo& 0.6757 & 0.7123 & 0.6368 & 0.6549\\
     & lv & 0.8995 & 0.8736& 0.9006 & 0.8998 \\
     & sn& -& -& -& - \\
     & so& 0.5551 & 0.5887 & 0.6577 & 0.5556 \\
     & sw& 0.8291 & 0.7981 & 0.8421 & 0.8125 \\
     & uz& 0.8621 & 0.8210 & 0.8608 & 0.8314\\
     \midrule
     \multirow{21}{*}{Unseen} & am& 0.2833 & 0.5560 & 0.2845 & 0.5018\\
     &bn&  0.8269  & 0.8791 & 0.9005 & 0.9430 \\
     &ca& 0.8733 & 0.8255 & 0.8750 & 0.8542 \\
     &es& 0.8518 & 0.8103 & 0.8583 & 0.8377 \\
     &fr& 0.8312 & 0.7607 & 0.8294 & 0.8447 \\
     &hi&0.7128 & 0.8210 & 0.8055 & 0.7981 \\
     &hr& 0.8531 & 0.8404 & 0.8532 & 0.8495 \\
     & ka&0.6289 & 0.8577 & 0.8103 & 0.8606 \\
     & ko& 0.5282& 0.8297& 0.7652 & 0.8381 \\
     &lij& 0.3319 & 0.2893 & 0.3333 & 0.2979 \\
     & my& 0.2128 & 0.5263 & 0.2785 & 0.5750 \\
     & or&  0.0708 & 0.4082 & 0.3851 & 0.2339 \\
     & pt& 0.8566 & 0.8015 & 0.8558 & 0.8449 \\
     & ro&0.8906 & 0.8548 & 0.8880 & 0.8768 \\
     & ru&0.7992 & 0.7922 & 0.8132 & 0.8051 \\
     & sq& 0.8658 & 0.8120 & 0.8627 & 0.8259 \\
     & sr&0.8540 & 0.8201 & 0.8790 & 0.8739\\
     & sv& 0.9075 & 0.8484 & 0.9076 & 0.8919 \\
     & te& 0.3278& 0.7441 & 0.5494 & 0.7632 \\
     & th& 0.5162 & 0.6841& 0.6320 & 0.6110 \\
     & ur& 0.8906 & 0.9208& 0.9205 & 0.9220 \\
     \bottomrule
    \end{tabular}%
    }
    \caption{F1 scores of each model trained on \textbf{dissim-same} language set. \textbf{Seen}: languages used during both pre-training. \textbf{Unseen}: languages not encountered during pre-training. Results are provided for four different input types: Orthographic (Ortho), IPA, Romanized (Rom), and Cipher (Cipher).}
    \label{tab:appx_dissimsame}
\end{table}

\begin{table}[]
\resizebox{\columnwidth}{!}{%
    \begin{tabular}{l|l|llll}
    \toprule
     & & Ortho   & IPA   & Rom   & Cipher \\
     \midrule
     
     \multirow{8}{*}{Seen} & am& 0.4941 & 0.5403&0.5760 &0.5364\\
     &bn& 0.9579 &0.9488 &0.9552 &0.9479\\
     &fr&  0.8528  &0.8265    & 0.8487  & 0.8432  \\
     & ka& 0.8647 & 0.8607   & 0.8619  & 0.8598 \\
     & ko& 0.7699 &  0.8347  & 0.8382  & 0.8333 \\
     & my&  0.5259 & 0.5738 & 0.5164  & 0.5477 \\
     & te& 0.7532 & 0.7529 & 0.7528  & 0.7734 \\
     & th& 0.7031 & 0.6813 &  0.6810 & 0.6727  \\
     \midrule
     \multirow{21}{*}{Unseen}
     & ca&0.8797 & 0.8503 & 0.8803 & 0.8513 \\
     &de&0.8088 & 0.7555 & 0.8134 & 0.7855 \\
     &es& 0.8615 & 0.8315 & 0.8687 & 0.8352\\
     &fi&  0.8504 & 0.8188 & 0.8532 &  0.8311 \\
     &hi& 0.6585 & 0.8223 & 0.8472   &  0.7939 \\
     &hr& 0.8642 & 0.8381 & 0.8652   &   0.8428   \\
     &ilo& 0.5272 & 0.5726 &  0.5122     &  0.4516    \\
     &lij&  0.3465 & 0.3243 &   0.3793 & 0.2833 \\
     & lv& 0.8948 & 0.8544   &  0.891 & 0.8762 \\
     & or&  0.3840 & 0.3931   &  0.4373 & 0.2913\\
     & pt&  0.8609 &  0.8245  & 0.8630 & 0.8437\\
     & ro&   0.8940 & 0.8746  & 0.8979  & 0.8788 \\
     & ru& 0.6753 & 0.7941   & 0.8207  & 0.8049  \\
     & sn&  -  & - & - & -  \\
     & so& 0.6140 &  0.4893  & 0.5462  & 0.5299 \\
     &sq& 0.8720 & 0.8395   & 0.8533   &  0.8389 \\
     & sr&  0.6697 & 0.8405  & 0.8826 & 0.8735\\
     & sv& 0.9117   & 0.8641   & 0.9119   &  0.8920 \\
     & sw& 0.7968 & 0.7527  &0.7855 & 0.7536\\
     & ur& 0.6974 & 0.9072& 0.9243 & 0.9226 \\
     & uz& 0.8317 & 0.8004   & 0.8300  & 0.8121\\
     \bottomrule
    \end{tabular}%
    }
    \caption{F1 scores of each model trained on \textbf{dissim-div} language set. \textbf{Seen}: languages used during both pre-training. \textbf{Unseen}: languages not encountered during pre-training. Results are provided for four different input types: Orthographic (Ortho), IPA, Romanized (Rom), and Cipher (Cipher).}
    \label{tab:appx_dissimdiff}
\end{table}

\begin{table*}[]
    \centering
    \resizebox{\linewidth}{!}{
    \begin{tabular}{ccccc|cccc|cccc|cccc}
    \toprule
        & \multicolumn{4}{c}{sim-same}& \multicolumn{4}{|c}{dissim-same}&\multicolumn{4}{|c}{sim-div}&\multicolumn{4}{|c}{dissim-div} \\ \midrule
        & Ortho & IPA & Rom & Cipher & Ortho & IPA & Rom & Cipher & Ortho & IPA & Rom & Cipher & Ortho & IPA & Rom & Cipher \\ \midrule
        de &0.6627&0.6389&0.6719&0.6529&0.6527&0.6513&0.6577&0.6613&0.6425&0.6391&0.6403&0.6483&0.6463&0.6469&0.6673&0.6543 \\ \midrule
        en &0.6822&0.6601&0.6864&0.6667&0.6647&0.6439&0.6814&0.6469&0.6739&0.6569&0.6834&0.6649&0.6930&0.6764&0.7072&0.7030 \\ \midrule
        es &0.6820&0.6691&0.6856&0.6725&0.6601&0.6563&0.6721&0.6487&0.6587&0.6579&0.6651&0.6571&0.6553&0.6603&0.6699&0.6615 \\ \midrule
        fr &0.6715&0.6519&0.6653&0.6591&0.6377&0.6355&0.6407&0.6353&0.6695&0.6433&0.6615&0.6567&0.6752&0.6649&0.6735& 0.6691\\ \midrule
        hi &0.4611&0.5790&0.6112&0.5890&0.5230&0.5802&0.6098&0.5747&0.6068&0.6036&0.6246&0.6174&0.3747&0.5970&0.6086&0.5962 \\ \midrule
        ru &0.6382&0.6181&0.6279&0.6094&0.6299&0.6146&0.6335&0.6228&0.6487&0.6531&0.6487&0.6455&0.3770&0.6279&0.6319&0.6407 \\ \midrule
        sw &0.5780&0.5894&0.6020&0.5727&0.5958&0.6100&0.5812&0.5978&0.5743&0.5749&0.5816&0.5868&0.5848&0.5848&0.5972&0.5890 \\ \midrule
        th &0.4032&0.4774&0.5148&0.5116&0.4323&0.5176&0.4752&0.4539&0.3826&0.5208&0.5118&0.5112&0.6210&0.5429&0.5609& 0.5661\\ \midrule
        tr &0.6299&0.6224&0.6397&0.6399&0.6343&0.6228&0.6176&0.6102&0.6142&0.6301&0.6437&0.6218&0.6337&0.6457&0.6295& 0.6102\\ \midrule
        ur &0.3970&0.5387&0.5615&0.5621&0.5271&0.5439&0.5609&0.5591&0.5717&0.5766&0.5745&0.5218&0.3501&0.5545&0.5701&0.5661 \\ \midrule
        vi &0.6257&0.6182&0.6375&0.6108&0.6429&0.6257&0.6405&0.5836&0.6132&0.6321&0.6371&0.6096&0.6263&0.6403&0.6431&0.6385 \\ 
        \bottomrule

    \end{tabular}
    }
    \caption{Accuracy of each model on sentence classification (i.e., XNLI) task.}
    \label{tab:xnli-accs}
\end{table*}

\begin{table*}[]
\centering
\resizebox{\linewidth}{!}{
\begin{tabular}{ccccc|cccc|cccc|cccc}
\toprule
 & \multicolumn{4}{c}{sim-same} & \multicolumn{4}{|c}{dissim-same} & \multicolumn{4}{|c}{sim-div} & \multicolumn{4}{|c}{dissim-div} \\ \midrule
 & Ortho & IPA & Rom & Cipher & Ortho & IPA & Rom & Cipher & Ortho & IPA & Rom & Cipher & Ortho & IPA & Rom & Cipher \\ \midrule

ca &0.9860&0.9680&0.9848&0.9830&0.9770&0.9514&0.9770&0.9743&0.9811&0.9575&0.9798&0.9763&0.9815&0.9591&0.9804&0.9770\\ \midrule
de &0.9480&0.9258&0.9475&0.9419&0.9465&0.9314&0.9475&0.9454&0.9445&0.9242&0.9452&0.9400&0.9453&0.9249&0.9454&0.9431\\ \midrule
es &0.9615&0.9569&0.9607&0.9584&0.9504&0.9390&0.9503&0.9468&0.9535&0.9464&0.9532&0.9506&0.9564&0.9479&0.9563&0.9519\\ \midrule
fi &0.9370&0.9194&0.9391&0.9335&0.9336&0.9218&0.9323&0.9302&0.9349&0.9215&0.9338&0.9320&0.9373&0.9260&0.9357&0.9375\\ \midrule
fr &0.9768&0.9636&0.9763&0.9747&0.9685&0.9516&0.9644&0.9629&0.9735&0.9646&0.9739&0.9715&0.9757&0.9684&0.9738&0.9736\\ \midrule
hi &0.8047&0.9432&0.9464&0.9448&0.7682&0.9376&0.9435&0.9365&0.9592&0.9593&0.9617&0.9585&0.5317&0.9476&0.9533&0.9464\\ \midrule
hr &0.9597&0.9476&0.9586&0.9528&0.9538&0.9362&0.9515&0.9427&0.9648&0.9594&0.9641&0.9606&0.9564&0.9488&0.9576&0.9542\\ \midrule
ka &0.6951&0.7912&0.8039&0.7916&0.5980&0.7658&0.7785&0.7687&0.4667&0.8021&0.8066&0.8008&0.8662&0.8567&0.8590&0.8568\\ \midrule
ko &0.4201&0.8551&0.8836&0.8687&0.4122&0.8400&0.8567&0.8601&0.4128&0.8665&0.8742&0.8712&0.8246&0.9047&0.9085&0.9158\\ \midrule
lv &0.9559&0.9411&0.9472&0.9457&0.9537&0.9423&0.9454&0.9441&0.9507&0.9430&0.9451&0.9417&0.9500&0.9442&0.9476&0.9432\\ \midrule
pt &0.9719&0.9515&0.9700&0.9679&0.9584&0.9234&0.9571&0.9528&0.9629&0.9321&0.9595&0.9561&0.9641&0.9349&0.9617&0.9579\\ \midrule
ro &0.9664&0.9607&0.9647&0.9619&0.9483&0.9447&0.9489&0.9434&0.9526&0.9540&0.9556&0.9477&0.9526&0.9539&0.9541&0.9509\\ \midrule
ru &0.9359&0.9126&0.9393&0.9314&0.9302&0.8988&0.9319&0.9147&0.9530&0.9415&0.9512&0.9497&0.6027&0.9169&0.9365&0.9309\\ \midrule
sq &0.8468&0.8317&0.8456&0.8322&0.6723&0.6809&0.6410&0.6229&0.7227&0.7288&0.7334&0.6989&0.6938&0.7291&0.6980&0.6534\\ \midrule
sv &0.9498&0.9379&0.9474&0.9457&0.9177&0.8926&0.9178&0.9050&0.9271&0.9135&0.9302&0.9200&0.9264&0.9138&0.9281&0.9219\\ \midrule
te &0.6428&0.8697&0.8712&0.8874&0.6346&0.8530&0.8773&0.8596&0.6273&0.8673&0.8778&0.8649&0.9032&0.9180&0.9011&0.8885\\ \midrule
th &0.4540&0.7830&0.7885&0.7817&0.4407&0.7676&0.7848&0.7563&0.2018&0.7927&0.7938&0.7810&0.8398&0.8146&0.8214&0.8134\\ \midrule
ur &0.8434&0.8732&0.8795&0.8776&0.8492&0.8487&0.8725&0.8689&0.9001&0.8984&0.9044&0.8961&0.4645&0.8741&0.8879&0.8797\\ \midrule
uz &0.7095&0.6894&0.7124&0.6985&0.7463&0.6737&0.7547&0.7455&0.6954&0.7172&0.6939&0.6924&0.6860&0.6987&0.7054&0.6791\\ 


\bottomrule
\end{tabular}
}
\caption{F1 scores of each model on the POS task under different language-set conditions.}
\label{tab:pos-f1}
\end{table*}

\end{document}